%% file: acl_latex.tex
\pdfoutput=1

\documentclass[11pt]{article}

\usepackage[]{acl}

\usepackage{times}
\usepackage{latexsym}

\usepackage[T1]{fontenc}

\usepackage[utf8]{inputenc}

\usepackage{microtype}

\usepackage{graphicx}
\usepackage{amsmath}
\usepackage{amsfonts}
\usepackage{booktabs}
\usepackage{soul}
\usepackage{float}
\usepackage{bm}
\usepackage{xspace}

\newcommand{\model}{\textsc{HIBRIDS}\xspace}
\newcommand{\data}{\textsc{GovReport-QS}\xspace}

\definecolor{selfcolor}{rgb}{0.99, 0.90, 0.84}
\definecolor{childcolor}{rgb}{0.87, 0.92, 0.97}
\definecolor{siblingcolor}{rgb}{0.89, 0.94, 0.85}
\definecolor{descendantcolor}{rgb}{1.0, 0.95, 0.81}

\definecolor{correctgreen}{rgb}{0.33, 0.51, 0.22}
\definecolor{incorrectred}{rgb}{0.75, 0.08, 0.08}

\definecolor{positiveblue}{rgb}{0.69, 0.88, 0.92}
\definecolor{negativeorange}{rgb}{0.98, 0.85, 0.65}

\definecolor{fullorange}{RGB}{236, 128, 60}

\newcommand{\hlc}[2][yellow]{{%
    \colorlet{foo}{#1}%
    \sethlcolor{foo}\hl{#2}}%
}

\title{\model: Attention with Hierarchical Biases \\for Structure-aware Long Document Summarization}

\author{Shuyang Cao \and Lu Wang \\
  Computer Science and Engineering \\
  University of Michigan \\
  Ann Arbor, MI \\
  \texttt{\{caoshuy, wangluxy\}@umich.edu}}

\begin{document}
\maketitle

\input{00_abstract}
\input{01_introduction}
\input{02_related_work}
\input{03_hierarchical_bias}
\input{04_qs_generation}

\input{05_experiment}
\input{06_result}
\input{07_analysis}

\input{08_conclusion}

\input{acknowledgements}

\input{ethical_consideration}

\bibliography{custom}
\bibliographystyle{acl_natbib}

\appendix
\input{appendix_A_annotation}
\input{appendix_B_data_detail}
\input{appendix_C_human_eval}
\input{appendix_D_visualization}
\input{appendix_E_sample_output}
\input{appendix_F_implementation}

\input{appendix_large_table_figure}

\end{document}

%% file: 00_abstract.tex
\begin{abstract}

Document structure is critical for efficient information consumption. However, it is challenging to encode it efficiently into the modern Transformer architecture. 
In this work, we present \model, which injects \underline{Hi}erarchical \underline{B}iases fo\underline{R} \underline{I}ncorporating \underline{D}ocument \underline{S}tructure into the calculation of attention scores. 
We further present a new task, hierarchical question-summary generation, for summarizing salient content in the source document into a hierarchy of questions and summaries, where each follow-up question inquires about the content of its parent question-summary pair. 
We also annotate a new dataset with $6,153$ question-summary hierarchies labeled on long government reports. 
Experiment results show that our model produces better question-summary hierarchies than comparisons on both hierarchy quality and content coverage, a finding also echoed by human judges. 
Additionally, our model improves the generation of long-form summaries from lengthy government reports and Wikipedia articles, as measured by ROUGE scores.

\end{abstract}

%% file: 01_introduction.tex
\section{Introduction}

Document structure facilitates information searching, reading comprehension, and knowledge acquisition by providing an informative overview of the content~\cite{10.2307/747765, 10.2307/747349, 10.2307/747358, https://doi.org/10.1002/tea.3660110307, 10.2307/44426154}. 
Specifically, for summarization, its utility is twofold: (1) \textit{Source} document structures, such as sections and paragraphs, can be instructive for summary generation~\cite{cohan-etal-2018-discourse,celikyilmaz-etal-2018-deep,zhang-etal-2019-hibert}; (2) Structures in \textit{output} summaries, e.g., timelines~\cite{shahaf2012trains,wang-etal-2015-socially} or aspects~\cite{angelidis-lapata-2018-summarizing}, can also ease content understanding.

Nonetheless, state-of-the-art abstractive summarization systems, all built on the Transformer architecture~\cite{zhang2020pegasus,lewis-etal-2020-bart}, use attentions to estimate relations between pairwise tokens and largely ignore document structures. While hierarchical encoding has been investigated~\cite{zhang-etal-2019-hibert,balachandran-etal-2021-structsum}, its need for training large amounts of additional parameters leads to increased memory footprint and thus limits the allowed input length. 
As for the output, the structure of single document summaries remains largely ``flat'', such as a list of aspects~\cite{meng-etal-2021-bringing}. We argue that it is imperative to develop systems that can output summaries with rich structures to support knowledge acquisition, which is especially critical for long documents that cover numerous subjects with varying details~\cite{huang-etal-2021-efficient, kryscinski2021booksum}.

\begin{figure*}[t]
    \centering
    \includegraphics[width=0.99\textwidth]{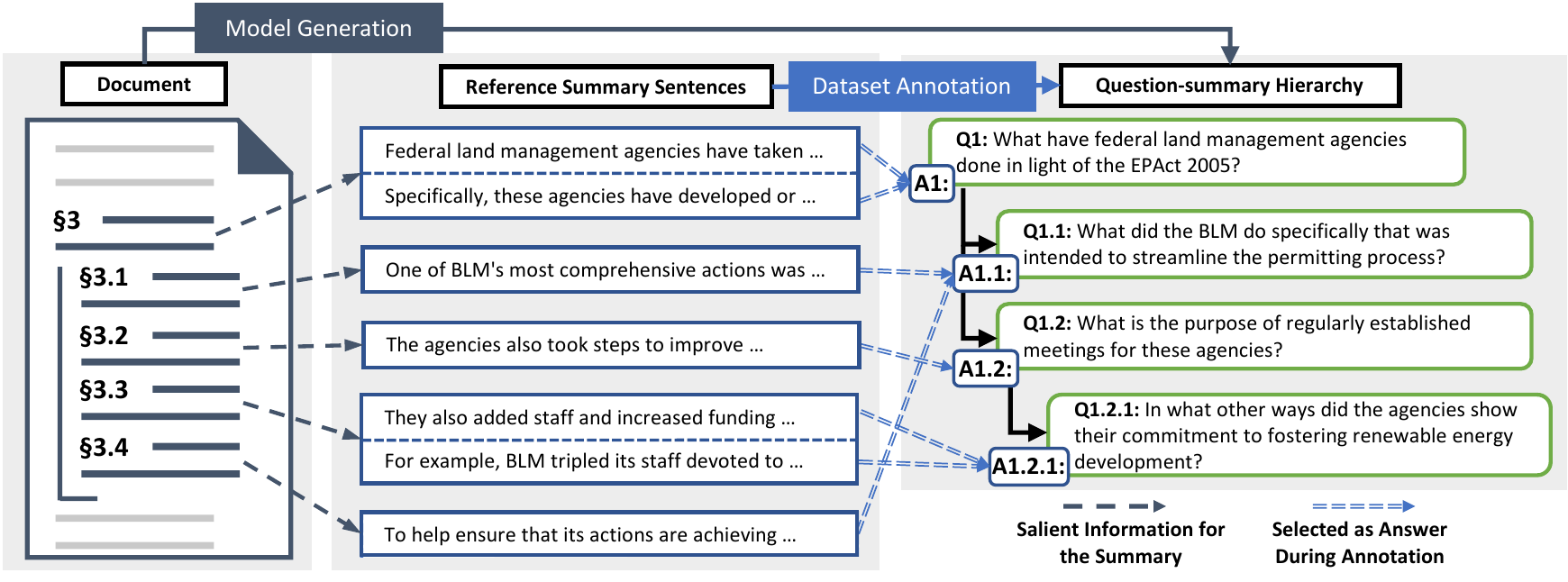}
    \caption{
    The question-summary hierarchy annotated for sentences in a reference summary paragraph. Summarization models are trained to generate the question-summary hierarchy \textit{from the document}, which signifies the importance of encoding the document structure. For instance, to generate the follow-up question-summary pairs of Q1.1 and A1.1 from A1, it requires the understanding of both the content and the parent-child and sibling relations among \S3, \S3.1, and \S3.4. 
    }
    \label{fig:intro_example}
\end{figure*}

This work consists of two main objectives: (1) effectively informing summarization models of the source document's structure, and (2) presenting a new summarization task that produces hierarchically organized question-summary pairs to facilitate information consumption. 
To this end, we propose \textbf{\model} (\underline{Hi}erarchical \underline{B}iases fo\underline{R} \underline{I}ncorporating \underline{D}ocument \underline{S}tructure).\footnote{Our code and newly collected data can be found at \url{https://shuyangcao.github.io/projects/structure_long_summ}.} 
We design learnable \textit{hierarchical biases}, as part of the Transformer attention calculation, to adjust attention weights based on tokens' relative positions with regard to the document structure, inspired by the relative position method that modifies attention calculation~\cite{JMLR:v21:20-074}. 
Concretely, we leverage the natural structure of a document, i.e., section levels, to construct a \textit{document structure tree} (Figure~\ref{fig:tree_structure}). 
Each learnable bias corresponds to the relation between a pair of sections, based on the distance between them in the structure tree. 
Intuitively, hierarchical biases adjust attention weights between tokens based on how conceptually close/distant their corresponding sections are, and they also enable summarizers to capture long-range relatedness for better document understanding.

Furthermore, we design a new summarization task, \textbf{hierarchical question-summary generation}: Given a document, automatically generate questions and summaries that are organized hierarchically to lay out details for topics at different levels. As shown in Figure~\ref{fig:intro_example}, 
each question asks about \textit{salient} content of the document (to be summarized) and its child questions focus on content in the corresponding summary.
%
This hierarchy not only exposes salient topics and their relations, but also allows readers to quickly identify aspects of interest to focus on. 
Our task design is inspired by the top-down knowledge learning process: People start by asking broad questions to acquire general knowledge, and then dive into details~\cite{hintikka1981logic,stede2004information}. 
Notably, as there is no available dataset with such annotations, we also label a new dataset, \textbf{\data}, consisting of $6{,}153$ question-summary (QS) hierarchies for summary paragraphs based on $1{,}714$ reports from the \textsc{GovReport} dataset~\cite{huang-etal-2021-efficient}. 
Each summary paragraph contains $4.07$ questions with an average QS hierarchy depth of $2.26$ levels.

We first compare \model with models that use structure-aware architectures~\cite{rohde2021hierarchical} and linear relative positions~\cite{JMLR:v21:20-074}. 
We conduct experiments on the hierarchical QS generation dataset using two setups: (1) generating a full hierarchy given the first question, and (2) generating follow-up questions given a QS pair. 
Automatic evaluation shows that our model produces better follow-up questions and summaries than comparisons, while also achieving better or comparable content coverage of full summaries, when compared with a hierarchical model~\cite{rohde2021hierarchical} that learns $2$M more parameters.
In human evaluation, \model is considered to build better hierarchies that require fewer manual corrections with more relevant summaries. 
We further test on the long document summarization task to produce full summaries using \textsc{GovReport} and \textit{a newly collected dataset} consisting of about $21$k high-quality biographies with summaries from Wikipedia. Again, our system summaries obtain uniformly higher ROUGE scores than comparisons, demonstrating the generalizability of \model. 

%% file: 02_related_work.tex
\section{Related Work}
\label{sec:related_work}

\paragraph{Document Structure-aware Summarization.}
Structural information has long been leveraged for identifying summary-worthy content, including discourse structures labeled by experts~\cite{marcu-1997-discourse} or automatic parsers~\cite{hirao-etal-2013-single, durrett-etal-2016-learning,xu-etal-2020-discourse}, and topical structures derived from lexical chains~\cite{barzilay1999using} or probabilistic models~\cite{barzilay-lee-2004-catching,daume2006bayesian}.
Natural structures of documents, such as sentences, have been used for pre-training a sentence-level encoder~\cite{zhang-etal-2019-hibert} or inducing dependencies among them~\cite{liu-etal-2019-single} for building extractive summarization systems.  
Based on separately encoded paragraphs, deep communication agents~\cite{celikyilmaz-etal-2018-deep} and inter-paragraph attentions~\cite{liu-lapata-2019-hierarchical} are employed to build abstractive summarization models by exchanging information from different paragraphs.
Using section structures, \citet{cohan-etal-2018-discourse} design a section-level encoder based on the output of a word-level encoder for long document summarization. 
Nevertheless, multi-level encoders are more expensive since they introduce a significant amount of parameters and add extra padding at multiple levels of model design. 
By contrast, \model effectively informs models of document structure by introducing a novel bias term in attention calculation among tokens, which only introduces a small number of learnable parameters.

\smallskip
\noindent \textbf{Long Document Summarization} also benefits from the inclusion of document structure information. For example, extractive summarization methods are developed to combine section-level and sentence-level information encoded by multi-level encoders~\cite{xiao-carenini-2019-extractive} and include longer context via sliding encoding over sections~\cite{cui-hu-2021-sliding}. 
Recent work on summarizing long documents focuses on designing efficient Transformers with sparse attentions to produce abstractive summaries for long documents in an end-to-end fashion~\cite{beltagy2020longformer,zaheer2020bigbird,huang-etal-2021-efficient}. 
However, they all ignore the natural structure of long documents, such as sections and subsections. 
Based on a simple design, \model can be integrated into any efficient Transformer seamlessly for incorporating document structure information.

\smallskip 
\noindent \textbf{Generating question-answer (QA) pairs} has been studied to facilitate information seeking within documents, mainly for producing questions that can be addressed by short phrases~\cite{du-cardie-2018-harvesting,10.1145/3366423.3380270}. Prior work mostly focuses on improving QA pair relevance by leveraging additional QA systems~\cite{sachan-xing-2018-self}, measuring roundtrip consistency~\cite{alberti-etal-2019-synthetic}, or refining questions iteratively~\cite{qu-etal-2021-asking}. 
Generating a two-level hierarchy of QA pairs from a given paragraph is investigated by \citet{krishna-iyyer-2019-generating}.
Our work is different in at least three aspects. 
First, our goal is to provide a structured summary that focuses on the \textit{salient content} of the given document, rather than creating questions about any generic information, as done in most QA data construction~\cite{rajpurkar-etal-2016-squad, choi-etal-2018-quac}. 
Second, our \data data concerns richer hierarchies for presenting content in long documents, e.g., $23.6\%$ of our hierarchies contain at least \textit{three levels}. 
Our parent-child pairs also cover diverse relations, e.g., adding explanations or expanding the topics, beyond asking about specific details as done in \citet{krishna-iyyer-2019-generating}. 
Third, our questions are designed to be open-ended and grounded in the given document, so our new task is more suitable for summarization models. 

%% file: 03_hierarchical_bias.tex
\section{\model with Hierarchical Biases}
\label{sec:model}

In this section, we first introduce how relative positions are defined over the document structure tree. 
Then we present \model, which can be included in encoder self-attentions or decoder cross-attentions to adjust the attention scores based on tokens' relative positions.

\begin{figure}[t]
    \centering
    \includegraphics[width=0.47\textwidth]{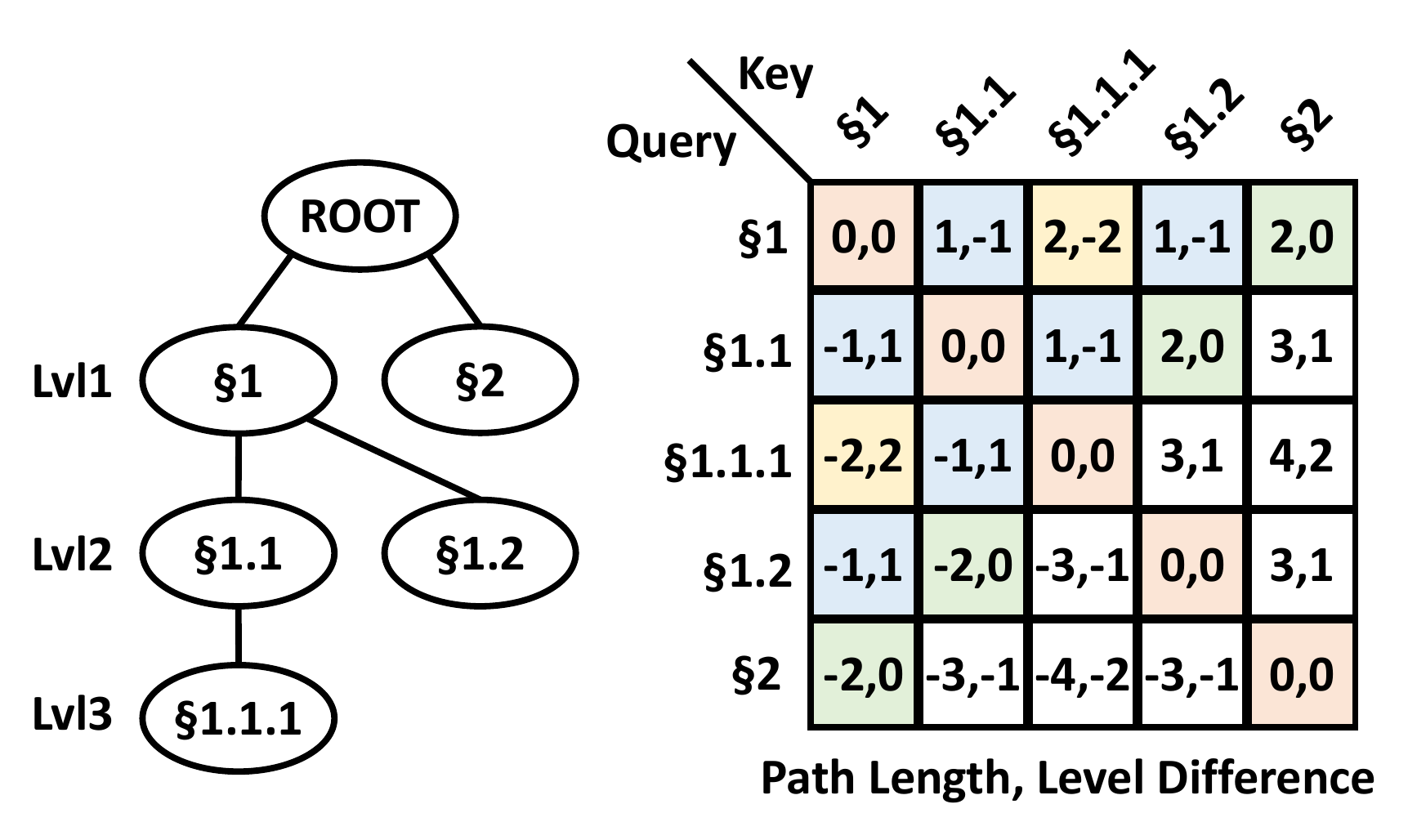}
    \caption{
    Example path lengths and level differences (right) that encode the relative positions with regard to the document tree structure (left). 
    Each query/key represents a block of tokens that belong to the same section. We highlight important section relations: \hlc[selfcolor]{self}, \hlc[childcolor]{parent-child}, \hlc[descendantcolor]{ancestor-descendant} (other than parent-child), and \hlc[siblingcolor]{sibling}. 
    From \S1 (level 1) to \S1.2 (level 2), the level difference is \textbf{-1} and path length is \textbf{1} since \S1 occurs \textit{before} \S1.2. 
    When looking back from \S1.2 to \S1, both numbers' signs are flipped, i.e, (\textbf{1}, \textbf{-1}). 
    }
    \label{fig:tree_structure}
\end{figure}

\subsection{Document Structure Tree and Tree-based Relative Positions}

We first construct a \textit{document structure tree} (Figure~\ref{fig:tree_structure}, left), by leveraging the natural structure of sections and subsections (henceforth sections) in documents, which is available in our experiment data extracted from government reports and Wikipedia articles. 
We then capture the relative position between pairwise tokens $x$ and $y$ in two different sections, e.g., $S_x$ and $S_y$, with two tree-based measures. 
(1) \texttt{PathLen}$(x, y)$: the length of the shortest path from $S_x$ to $S_y$; 
(2) \texttt{LvlDiff}$(x, y)$: the level difference from $S_x$ to $S_y$. 
\texttt{PathLen} is designed to be \textit{asymmetric} to capture content ordering, i.e., its value is positive if $S_x$ appears before $S_y$ in the document, and vice versa. 
Examples are displayed in Figure~\ref{fig:tree_structure}. 

\subsection{Attentions with Hierarchical Biases} 
The design of \model is based on a lookup table $B[\cdot, \cdot]$: Each item in it corresponds to a learnable hierarchical bias defined by path length and level difference, which is then used to bias the attention calculation for tokens in different sections. Each head maintains its own lookup table $B$. 

We first apply \model to Transformer encoder self-attention computation, which is called \textbf{\model-\textsc{enc}}. Given the $i$-th query $\bm{q}_i$ and the matrix $\bm{K}$ formed by $n$ keys for all input tokens, 
\model adds a bias for each key, with respect to the $i$-th query, to attention calculation:

{
    \fontsize{10}{11}\selectfont
    \begin{equation}
        a_{ij} = \mathrm{softmax} (\bm{q}_i \bm{K}^T + \bm{b}_i)_j
        \label{eq:attn_bias}
    \end{equation}
}%
where the vector $\bm{b}_i = [ b_{i1}, \dots, b_{ij}, \dots, b_{in} ]$ contains the bias terms derived from our hierarchical biases as follows:

{
    \fontsize{10}{11}\selectfont
    \begin{equation}
    b_{ij} = B[\mathtt{PathLen}(i, j), \mathtt{LvlDiff}(i, j)]
    \label{eq:enc_bias}
    \end{equation}
}%
where $\mathtt{PathLen}(i, j)$ and $\mathtt{LvlDiff}(i, j)$ are the path length and level difference between the sections that tokens $i$ and $j$ belong to. Note that $b_{ij}$ varies among different heads. 
\model-\textsc{enc} guides tokens to attend to structurally related tokens during encoding. 

We then apply \model to decoder cross-attention calculation, named as \textbf{\model-\textsc{dec}}, to encourage more coherent generation by establishing better alignment with the source document. 
At the generation step $t$, the cross-attention weight to the $j$-th input token adjusted by bias $b_{tj}$ is obtained similarly as in Eq.~\ref{eq:attn_bias} with the following modification. 
%
%
We calculate $b_{tj}$ as the weighted sum of the hierarchical biases for all input tokens (indexed with $l$) to the $j$-th token. The weight is chosen as the decoder's second last layer's cross-attention score between the $t$-th generated token and the $l$-th input token, which is shown to better capture word alignment~\cite{garg-etal-2019-jointly,cao-wang-2021-attention}. 
$b_{tj}$ is only applied to the decoder's last layer with the following formulation:

{
    \fontsize{10}{11}\selectfont
    \begin{equation}
        b_{tj} = \sum_{l} a^{crs}_{tl} \cdot B[\mathtt{PathLen}(l, j), \mathtt{LvlDiff}(l, j)] 
    \end{equation}
}%
where $a^{crs}_{tl}$ is the decoder's second last layer's cross-attention weight for the generation step $t$ to the $l$-th input token.

\paragraph{\textsc{HIBRIDS}$_{\textsc{s}}$ with Selected Relations.} 
We further consider only keeping salient relations from the tree to reduce the number of parameters to learn, including \textit{self} (same section), \textit{parent-child}, \textit{ancestor-descendant}, \textit{sibling}, \textit{neighboring in text}, 
and \textit{within the same top-level section} (e.g., \S1.1.1 and \S1.2 are both in \S1). 
In total, they account for $21.6\%$ of all relation occurrences. 
The modified \textsc{HIBRIDS}$_{\textsc{s}}$ can also be applied to both encoder and decoder. 

%% file: 04_qs_generation.tex
\section{A New Task: Hierarchical Question-summary Generation}
\label{sec:hier_qs_gen}

We introduce a new summarization task in this section: Given a document or several sections of a document, we aim to generate question-summary (QS) pairs that are organized hierarchically. As shown in Figure~\ref{fig:intro_example}, this QS hierarchy lays out details for topics at multiple levels, with each child QS pair expanding the content of its parent. 
Our task is motivated by how human learns knowledge in a top-down fashion, where general knowledge is acquired first and details and in-depth content are explored later~\cite{hintikka1981logic}. This hierarchy proactively highlights the document structure, to further promote content engagement and comprehension~\cite{10.2307/25655454}. 

\subsection{Question-summary Hierarchy Annotation Procedure} 

We first annotate a new dataset, \data, with hierarchical QS pairs, based on articles and corresponding summaries selected from the \textsc{GovReport} dataset~\cite{huang-etal-2021-efficient}. As these documents and summaries have $9{,}409$ and $553$ words on average respectively, directly annotating full documents with a QS hierarchy presents a challenge. 
To address this, we ask annotators to create hierarchical questions for a selected summary paragraph and only allow them to select complete sentences from the summary paragraph as the corresponding answers. 
Each question created should be fully addressed by its answer and the answer should not contain information irrelevant to the question. 
For follow-up questions, they are encouraged to ask about specific details or issue questions that can yield summaries that elaborate from their parents.
Annotators are also instructed to construct hierarchies of as many levels as possible. 
Figure~\ref{fig:intro_example} demonstrates how hierarchical questions are created and how answer sentences are selected when annotating a report on the development of renewable energy. 

To cover more documents and avoid collecting shallow hierarchies, each summary paragraph is annotated by one annotator and we select high-quality summary paragraphs for annotation based on heuristic rules, e.g., each paragraph should have at least $3$ sentences and $70$ words and an adequate level of abstractiveness as measured by normalized density of extractive fragments~\cite{grusky-etal-2018-newsroom} (with a threshold of $< 0.15$). 
Annotation instructions and details of paragraph selection are in Appendix~\ref{appendix:annotation}. 

We hired 11 college students who are native English speakers to carry out the annotation tasks in multiple rounds. Feedback was provided to each annotator after each round. A finalization stage was conducted after collecting all annotations, where 4 high-quality annotators were asked to correct typos, remove factoid questions, and make minor adjustment to the hierarchies when errors were detected. 

\paragraph{\data Statistics.}
In total, $6{,}153$ summary paragraphs are annotated with $25{,}055$ QS pairs. 
On average, $4.07$ QS pairs are created per summary paragraph, spanning $2.26$ levels. 
$70.5\%$ and $23.6\%$ of paragraphs are annotated with \textit{two} and \textit{three} levels of questions, making our dataset a valuable benchmark for studying QS hierarchy generation, query-focused summarization, and question generation.

\subsection{Aligning Summary Paragraphs with Document Sections} 
The QS hierarchies then become the target generation, and we construct inputs to our QS hierarchy generation system by mapping annotated summary paragraphs back to \textit{sections} in source documents. 

Concretely, we match each summary sentence to a document paragraph based on a combination of BERT-based, word overlap-based, and entity overlap-based similarities (details in Appendix~\ref{appendix:annotation}). 
All \textit{sections} where matched paragraphs belong, along with the titles of their ancestor sections, are combined together to serve as the system input for generating the corresponding QS hierarchy, as demonstrated in Figure~\ref{fig:intro_example}. 
The paired sections have an average length of $2{,}029$, longer than documents in many standard summarization benchmarks.

%% file: 05_experiment.tex
\section{Experiment Setups}
\label{sec:experiments}

\subsection{Datasets and Tasks} 

We evaluate \model on three different tasks with outputs of varying structures. 

\paragraph{Task I: QSGen-Hier.} 
Based on \data, we first experiment with a setup where, given the aligned document sections and a root question, the model is expected to produce a summary that addresses the question as well as the rest of the hierarchy. 
To linearize a QS hierarchy for the Transformer sequential decoder, we concatenate its QS pairs following a depth-first traversal. Special tokens are inserted before each QS pair to indicate the change of its level from the previous QS pair: \texttt{[L$\downarrow$]}, \texttt{[L$\uparrow$]}, and \texttt{[L-]} indicate that the level has incremented, decremented, and not changed, respectively. 
For example, the sample hierarchy in Figure~\ref{fig:intro_example} can be formulated as: ``\textit{A1} \texttt{[L$\downarrow$]} \textit{Q1.1} \textit{A1.1} \texttt{[L-]} \textit{Q1.2} \textit{A1.2} \texttt{[L$\downarrow$]} \textit{Q1.2.1} \textit{A1.2.1}''. 
On this task, we divide our samples into train/dev/test splits with sizes of $4{,}878/{644}/{631}$.

\paragraph{Task II: QSGen-ChildQ.} 
Next, we leverage \data for follow-up question generation: Given a QS pair and the aligned document sections, we aim to generate all child questions. 
With this setup, two samples can be created from the example in Figure~\ref{fig:intro_example}. 
The first one takes as input \textit{``Q1 A1''} and the aligned sections to generate \textit{``Q1.1 Q1.2''}, whereas the other reads in \textit{``Q1.2 A1.2''} and the aligned sections to produce \textit{``Q1.2.1''}.
Here we construct train/dev/test splits with sizes of $7{,}157/{958}/{942}$.

\paragraph{Task III: Full Summary Generation.} 
We also conduct experiments on \textsc{GovReport} to test \model on generating long-form summaries for long inputs. We use the original data splits with $17{,}516/974/973$ samples in train/dev/test sets. 
We further collect \textit{a new dataset} from WikiProject Biography\footnote{\url{https://en.wikipedia.org/wiki/Wikipedia:WikiProject_Biography}} (\textsc{WikiBioSum}) to perform biography summarization. 
After collecting all available biographies, we keep the ones with at least two levels of section hierarchy and preserve section structures of all levels. For each article, the paragraph before the first section is treated as the target summary, and the rest becomes the input. 
The finalized dataset has $20{,}833$ pairs, divided into $18{,}751/1{,}041/1{,}041$ samples for train/dev/test sets. The average lengths of the input and output for \textsc{WikiBioSum} are $3{,}478$ and $1{,}266$.
Details of \textsc{WikiBioSum} data collection and filtering procedures are in Appendix~\ref{appendix:dataset_detail}. 

We set the maximum input length to $5{,}120$ for QSGen-Hier, QSGen-ChildQ, and full document summarization on \textsc{WikiBioSum}. On \textsc{GovReport}, the limit is set to $16{,}384$.

\subsection{Evaluation and Comparisons}

\paragraph{Evaluation Metrics.} 
We use ROUGE~\cite{lin-2004-rouge} for summarization evaluation and additionally report BLEU up to 4-gram~\cite{papineni-etal-2002-bleu} for evaluating the generated questions. 

We propose to evaluate the generated QS hierarchy against the reference hierarchy with F1 scores calculated as follows, inspired by labeled attachment score in dependency parsing~\cite{zeman-etal-2017-conll}:
We first map each generated QS pair to a reference QS pair following the highest sum of ROUGE-1 and ROUGE-2 scores between their summaries. 
%
After that, we consider two QS pairs with parent-child relation in the generated hierarchy. A \textit{match} is established only when their mapped QS pairs have a parent-child or ancestor-descendant relation in the reference hierarchy. Precision can then be calculated based on the matching results. We further weight each match based on the sum of the ROUGE-1 and ROUGE-2 scores calculated over both parent and child summaries. Weighted recall and F1 are calculated similarly.

\paragraph{Comparisons.} 
All tasks in this work involve long inputs. To allow efficient encoding, we use \textbf{\textsc{Longformer}}~\cite{beltagy2020longformer} with a window size of $1024$ as the base model, and fine-tune it for all systems and comparisons.

We first consider comparisons by adding special tokens to encode document structure: 
(1) \textbf{\textsc{SecTok}} inserts a special token \texttt{[SEC]} at the start of each section.
(2) \textbf{\textsc{LvlSecTok}} further differentiates sections at varying levels using different tokens (e.g., \texttt{[SEC-L1]} for \S1, \texttt{[SEC-L2]} for \S1.1).

Based on \textsc{LvlSecTok}, we build all \model variants and other comparisons listed below:

$\bullet$ \textbf{\textsc{HierEnc}}: We implement the hierarchical model by \citet{rohde2021hierarchical}, where we replace its sentence encoder with a section encoder of $12$ layers to maintain section structures. Among all models, \textsc{HierEnc} requires the most architecture change and adds the most parameters to learn.  

$\bullet$ \textbf{\textsc{MultiTask}}: We also consider predicting the selected relations used by \textsc{HIBRIDS}$_{\textsc{s}}$ (\S\ref{sec:model}) in a multi-task prediction setup with a bilinear classifier, operating on the representations of section tokens. We use equal weights for prediction loss and summarization loss.

$\bullet$ \textbf{\textsc{TokBias}} uses linear relative position biases as in T5~\cite{JMLR:v21:20-074}, which changes Eq.~\ref{eq:enc_bias} to $b_{ij} = R[i-j]$ where $R[\cdot]$ is a lookup table with each item corresponding to a learnable bias for a given relative distance. 

$\bullet$ \textbf{\textsc{SecBias}} replaces token-level linear distance in \textsc{TokBias} with section-level linear distance. 

Notably, \textsc{Longformer} and models using special tokens have $4.59$M parameters. 
\model and models with linear relative position biases use about $4.60$M parameters in total. On the other hand, \textsc{HierEnc} and \textsc{MultiTask} modify the architecture and have $6.62$M and $4.66$M parameters, which is less efficient for learning compared with models that use bias terms to adjust attention calculation. 

\begin{table}[t]
    \centering
    \small
    \setlength{\tabcolsep}{4.5pt}
    \begin{tabular}{lccccc}
    \toprule
        & \textbf{Hier} & \multicolumn{3}{c}{\textbf{Summary}} & \textbf{Ques} \\
        \textbf{Model} & \textbf{F1} & \textbf{R1} & \textbf{R2} & \textbf{RL} & \textbf{B4} \\
        \midrule
        \textsc{Longformer} & 12.67 & 42.34 & 16.18 & 37.60 & 10.00 \\
        \textsc{SecTok} & 12.86 & 42.67 & 16.34 & 38.01 & 10.02 \\
        \textsc{LvlSecTok} & 12.74 & 42.34 & 16.31 & 37.61 & 10.09 \\
        \midrule
        \multicolumn{6}{l}{\textit{Structure-aware Comparisons}} \\
        \textsc{HierEnc} & 11.77 & \textbf{42.82} & 16.32 & \textbf{38.06} & 9.89 \\
        \textsc{MultiTask} & 12.64 & 41.19 & 15.49 & 36.58 & 9.66 \\
        \midrule
        \multicolumn{6}{l}{\textit{Models with Linear Bias}} \\
        \textsc{TokBias} & 12.43 & 42.58 & 16.41 & 37.71 & 10.06 \\
        \textsc{SecBias} & 12.54 & 42.54 & 16.39 & 37.80 & 10.00 \\
        \midrule
        \multicolumn{6}{l}{\textit{Our Models}} \\
        \model-\textsc{enc} & \textbf{13.26} & 42.74 & \textbf{16.55} & 38.03 & \textbf{10.16} \\
        \textsc{HIBRIDS}$_{\textsc{S}}$-\textsc{enc} & 13.16 & 42.50 & 16.16 & 37.69 & 10.09 \\
        \model-\textsc{dec} & 12.68 & 42.31 & 16.17 & 37.58 & 9.75 \\
        \textsc{HIBRIDS}$_{\textsc{S}}$-\textsc{dec} & 12.71 & 42.44 & 16.42 & 37.82 & 9.84 \\
        \bottomrule
    \end{tabular}
    \caption{
    Results for QSGen-Hier on \data. The best result per metric is \textbf{bolded}. 
    Applying \model on the encoder produces better QS hierarchies (higher F1) and questions (higher BLEU). 
    Our models also yield better or comparable ROUGE scores, especially compared with \textsc{HierEnc} which requires 43\% more parameters and extra engineering efforts for architecture change. 
    \textbf{Ques}: question; \textbf{Hier}: hierarchy.
    }
    \label{tab:qsgen_fq_result}
\end{table}

%% file: 06_result.tex
\section{Experiment Results}

\subsection{Hierarchical Question-summary Generation}
\label{sec:qs_result}

\begin{figure}[t]
    \centering
    \includegraphics[width=0.48\textwidth]{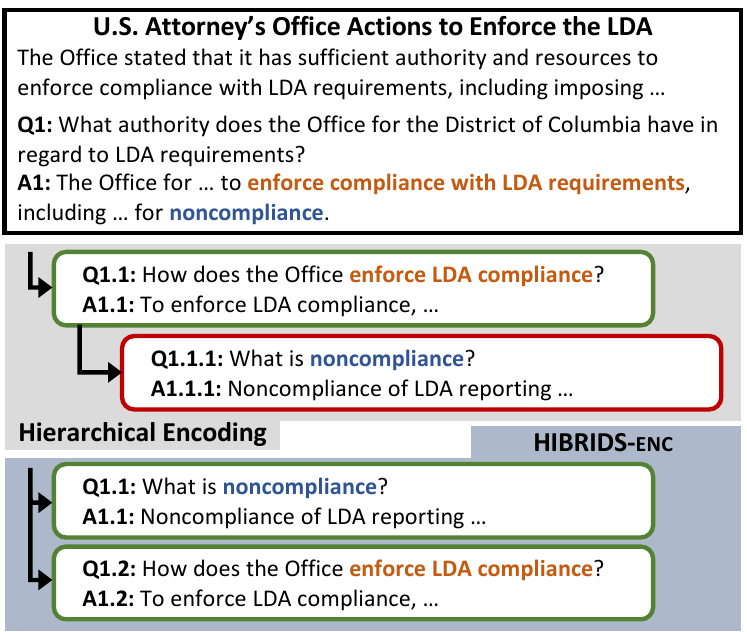}
    \caption{
    Sample output by the hierarchical encoding model (\textsc{HierEnc}) and \model-\textsc{enc}. 
    Our generated structure makes more sense with the constructed follow-up questions to Q1, highlighted in \textcolor{correctgreen}{\textbf{green}}, than the comparison model \textsc{HierEnc}. 
    }
    \label{fig:result_example}
\end{figure}

\paragraph{Results on QSGen-Hier.} 
We report results on the task of generating QS hierarchies in Table~\ref{tab:qsgen_fq_result}. 
\textit{\model-\textsc{enc} uniformly outperforms other variants and all comparisons on all metrics,} except for ROUGE-1 and ROUGE-L scores by \textsc{HierEnc}.
Note that \textsc{HierEnc} learns $2$M more new parameters than our models, and it produces QS hierarchies of lower quality despite its competitive ROUGE scores (Figure~\ref{fig:result_example}). 
This signifies the effectiveness of our design that directly injects structural information into word-level relation computation.
Meanwhile, \textit{\model on encoder is better at hierarchy quality than its variant on decoder}, suggesting the importance of resolving section relations during encoding.

Though not reported here, we experiment with \model on both the encoder and the decoder, and it results in degraded performance. One possible cause is that \model functions differently in these two setups (discussed in \S\ref{sec:additional_study}). We will explore better fusion techniques in future work.

\paragraph{Results on QSGen-ChildQ.}
Results on generating follow-up questions further validate the usefulness of hierarchical biases as shown in Table~\ref{tab:qsgen_childq_result}, where \textit{questions generated by \model-\textsc{enc} have the best quality} as measured by all metrics except for BLEU. \textsc{SecBias}, which is aware of section-level linear distance, also obtains outstanding performance, since it focuses on intra-section information and thus better determines what child questions should be asked for better relevance. 

\begin{table}[t]
    \centering
    \small
    \setlength{\tabcolsep}{4.5pt}
    \begin{tabular}{lcccc}
    \toprule
        \textbf{Model} & \textbf{R1} & \textbf{R2} & \textbf{RL} & \textbf{B4} \\
        \midrule
        \textsc{Longformer} & 26.90 & 8.69 & 25.57 & 14.44 \\
        \textsc{SecTok} & 26.76 & 8.82 & 25.42 & 14.51 \\
        \textsc{LvlSecTok} & 26.80 & 8.75 & 25.52 & 14.33 \\
        \midrule
        \multicolumn{5}{l}{\textit{Structure-aware Comparisons}} \\
        \textsc{HierEnc} & 26.38 & 8.81 & 24.99 & 14.54 \\
        \textsc{MultiTask} & 26.84 & 8.46 & 25.41 & 14.59 \\
        \midrule
        \multicolumn{5}{l}{\textit{Models with Linear Bias}} \\
        \textsc{TokBias} & 26.73 & 8.69 & 25.38 & 14.43 \\
        \textsc{SecBias} & 27.25 & 9.07 & 25.92 & \textbf{14.76} \\
        \midrule
        \multicolumn{5}{l}{\textit{Our Models}} \\
        \model-\textsc{enc} & \textbf{27.33} & \textbf{9.46} & \textbf{26.00} & 14.73 \\
        \textsc{HIBRIDS}$_{\textsc{S}}$-\textsc{enc} & 26.41 & 8.74 & 24.99 & 14.44 \\
        \model-\textsc{dec} &  27.17 & 8.67 & 25.71 & 14.36 \\
        \textsc{HIBRIDS}$_{\textsc{S}}$-\textsc{dec} &  26.29 & 8.50 & 25.09 & 14.30 \\
        \bottomrule
    \end{tabular}
    \caption{
    Results for QSGen-ChildQ. The best result per metric is \textbf{bolded}. 
    Using \model on encoder generates better follow-up questions according to ROUGE scores. 
    }
    \label{tab:qsgen_childq_result}
\end{table}

\smallskip
\noindent \textbf{Human evaluation} is conducted on QSGen-Hier, for five models with the highest automatic scores, to help understand how well the generated hierarchies are structured. 
We hire three judges who have \textit{extensive experience} in summarization annotation and evaluation tasks to assess $50$ groups of question-summary hierarchies. 
Human inspection on randomly selected outputs shows that most system generations have an appropriate coverage of the salient content in the source. 
Therefore, we focus on evaluating both global coherence and local coherence of the QS hierarchies based on the following two aspects. 
First, we ask evaluators to correct each generated hierarchy by rearranging the QS pairs so that each pair is attached to the parent that forms the best follow-up relation in steps. 
For each step, they are only allowed to attach a pair to its grandparent or sibling (i.e., the parent or child of its current parent).
They then report the \textbf{number of edits} conducted for the rearrangement. 
Second, for each QS pair, we ask them to determine if the question can be \textbf{answered} by the summary. 
Details of human evaluation are in Appendix~\ref{appendix:human_eval}.

As can be seen from Table~\ref{tab:qsgen_fq_human}, QS hierarchies generated by \model-\textsc{enc} model contain the best structured summaries as they require the fewest number of corrections and the generated questions are also more likely to be addressed by the corresponding summaries. Despite being competitive on automatic metrics, \textsc{SecTok} generates hierarchies that require the most corrections. 
Upon additional inspection, we find that \model's outputs often have better local coherence than the comparisons. 
Additionally, all models struggle to generate more engaging questions, which poses another challenge to future studies.

\begin{table}[t]
    \centering
    \small
    \setlength{\tabcolsep}{2pt}
    \begin{tabular}{lccc}
    \toprule
        \textbf{Model} & \textbf{\# of Edits} ($\downarrow$) & \textbf{Answerable Qs} ($\uparrow$) \\
        \midrule
        \textsc{SecTok} & 4.73 & 81.8\% \\
        \textsc{LvlSecTok} & 4.62 & 78.6\%   \\
        \textsc{HierEnc} & 4.17 & 81.4\%  \\
        \textsc{TokBias} & 3.77 & 82.8\%  \\
        \model-\textsc{enc} & \textbf{3.67} & \textbf{84.1\%} \\
        \bottomrule
    \end{tabular}
    \caption{Human evaluation results on QSGen-Hier. Hierarchies produced by \model-\textsc{enc} require fewer correction edits by human and contain more answerable questions by the generated summaries. 
    Krippendorff’s $\alpha$: $0.55$, $0.44$.}
    \label{tab:qsgen_fq_human}
\end{table}

\subsection{Full Summary Generation}
\label{sec:full_summ_result}

\begin{figure}[t]
    \centering
    \includegraphics[width=0.47\textwidth]{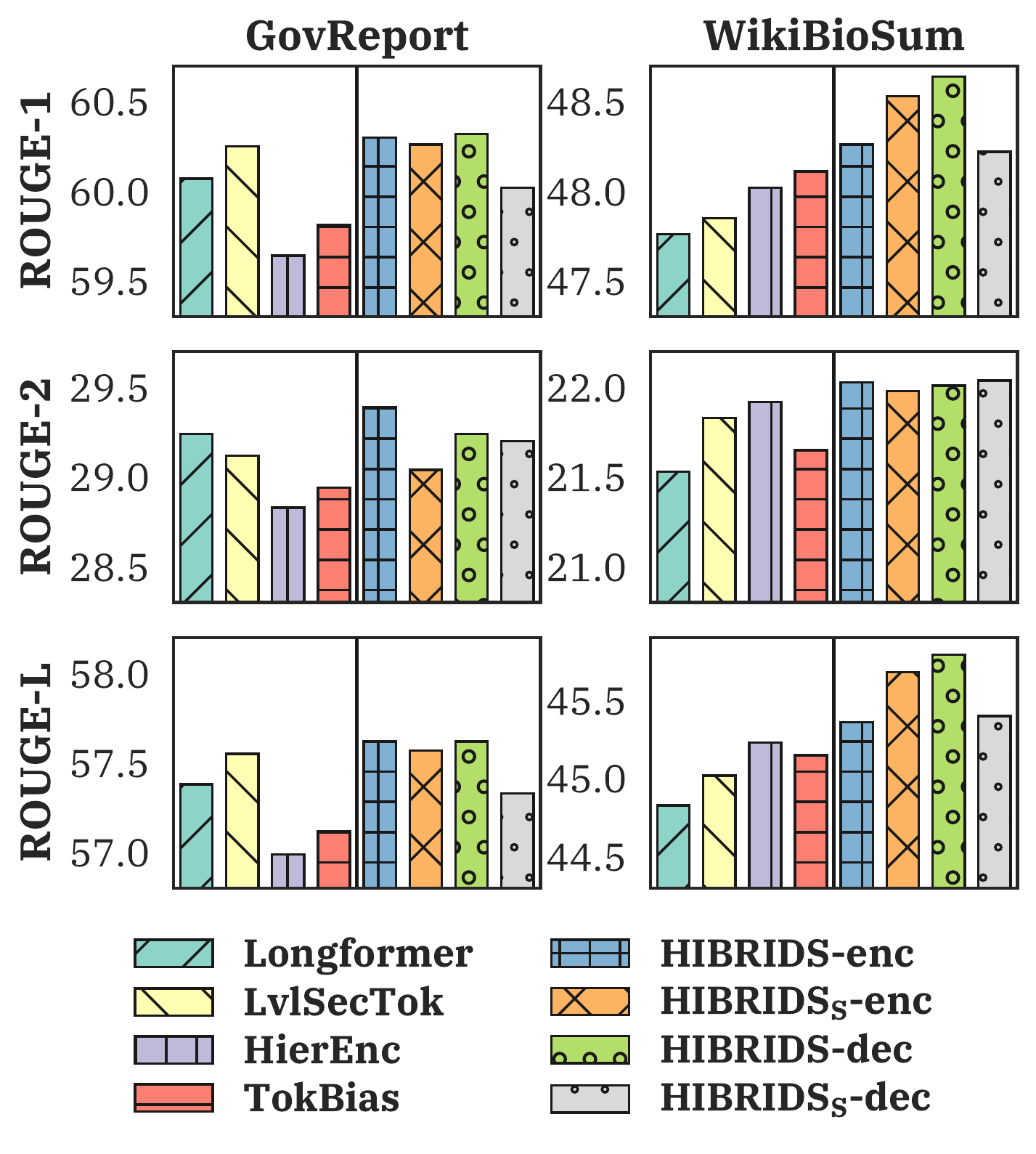}
    \caption{Results on full summary generation.  In each subfigure, the left panel includes models for comparisons and the right panel shows our models. 
    \model on either encoder and decoder uniformly outperforms the comparisons on both datasets.
    }
    \label{fig:summary_result}
\end{figure}

As demonstrated in Figure~\ref{fig:summary_result}, \textit{\model with full hierarchical biases outperform all comparisons} on both datasets, suggesting that our design of including structural relations in bias terms can generalize to other tasks. 
Compared to the results on QS hierarchy generation, using \model on the decoder yields greater improvement on full summary generation, especially in the biography domain where \model-\textsc{dec} obtains the best performance. It is likely that the longer summary length and higher compression ratio on \textsc{WikiBioSum} ($1{,}266$ and $0.45$) makes generation coherence more important by using better alignment. 
This highlights how hierarchical biases can aid long text generation. 

%% file: 07_analysis.tex
\section{Further Analyses}
\label{sec:additional_study}

\subsection{Visualizing the Learned Biases}
Here we aim to understand what is learned by our hierarchical biases. 
For \model-\textsc{enc} and \model-\textsc{dec} trained on QSGen-Hier, we visualize the values of their \textit{learned} hierarchical biases averaged over all heads at all layers for each (path length, level difference) pair on an example structure. 
Additional visualization is in Appendix~\ref{appendix:visualization}. 

From Figure~\ref{fig:bias_value} we see that using \model on the encoder encourages models to encode various relations, e.g., by upweighing grandparent (\S1.1.1 to \S1, \S1.1.1.1 to \S1.1) and preceding sibling (\S1.2 to \S1.1), and downweighing children (\S1 to \S1.1 and \S1.2, \S1.1 to \S1.1.1). 
This highlights the need of learning heterogeneous relations among sections beyond token distances. 
By contrast, \model on the decoder consistently biases towards parent and sibling contexts.
It might be because that the generation of fluent and coherent question-summary pairs relies on being aware of the scope of sections at the same or higher levels.

\begin{figure}
    \centering
    \includegraphics[width=0.48\textwidth]{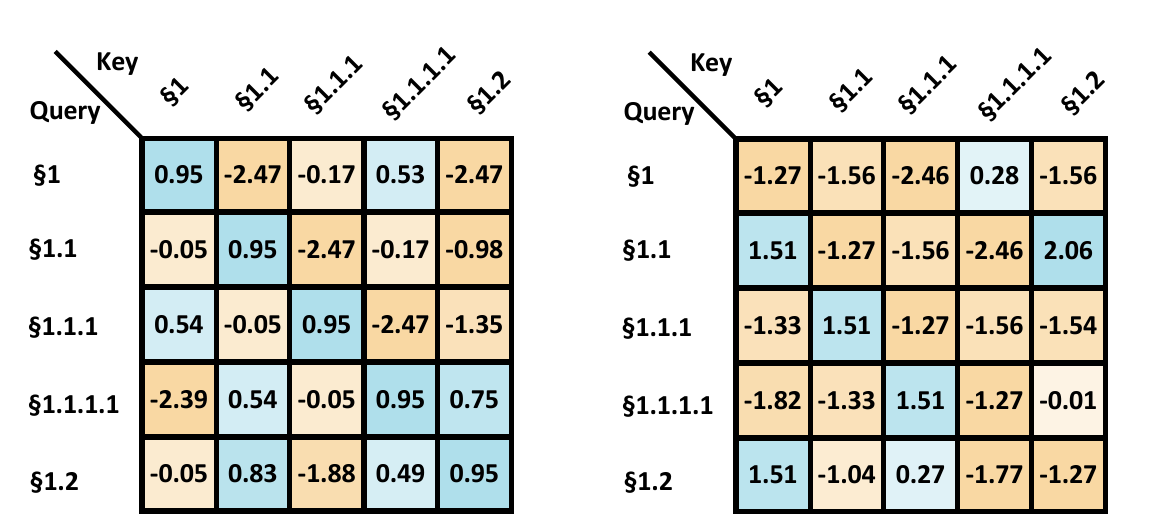}
    \caption{Visualization of hierarchical biases in \model-\textsc{enc} (left) and \model-\textsc{dec} (right) on QSGen-Hier. Positive and negative values are shaded in \hlc[positiveblue!49]{\textbf{blue}} and \hlc[negativeorange!49]{\textbf{orange}}. Displayed values are $100$X of actual values. 
    \model-\textsc{enc} biases towards current, grandparent and preceding sibling sections, while \model-\textsc{dec} focuses on parent and succeeding sibling sections. 
    }
    \label{fig:bias_value}
\end{figure}

\subsection{Ablation Study for \model}
We examine which design choices contribute the most to the performance gain by \model, by carrying out ablation studies on QSGen-Hier with \model-\textsc{enc}. 
We consider taking out (1) level difference, (2) path length, and (3) asymmetry of path length.  
As shown in Table~\ref{tab:ablation_result}, removing any component reduces summaries' content coverage and hierarchy quality, underscoring their contributions in more precisely representing structural relations for better document encoding. 
Level difference adds the most to hierarchy quality, as levels directly signal when to generate follow-up questions. 

\begin{table}[t]
    \centering
    \small
    \setlength{\tabcolsep}{3pt}
    \begin{tabular}{lccc}
    \toprule
        & \textbf{Summary} & \textbf{Question} & \textbf{Hierarchy} \\
        \textbf{Model} & \textbf{RL} & \textbf{B4} & \textbf{F1} \\
        \midrule
        \model-\textsc{enc} & 38.03 & 10.16 & 13.26 \\
        w/o Level Difference & \hlc[fullorange!40]{--0.50} & \hlc[fullorange!8]{--0.08} & \hlc[fullorange!40]{--0.51} \\
        w/o Path Length & \hlc[fullorange!30]{--0.43} & +0.05 & \hlc[fullorange!16]{--0.18} \\
        w/o Asymmetric Path & \hlc[fullorange!15]{--0.15} & \hlc[fullorange!12]{--0.12} & \hlc[fullorange!16]{--0.18} \\
        \bottomrule
    \end{tabular}
    \caption{Ablation study results. Performance change compared to the full model are reported. 
    Larger decreases of metrics are shaded with darker \hlc[fullorange!30]{orange}. 
    Removing level difference hurts the hierarchy quality substantially. 
    }
    \label{tab:ablation_result}
\end{table}

\subsection{Can \model Improve Hierarchical Encoding?}
We further study if \model can boost the section encoder of \textsc{HierEnc}.
Table~\ref{tab:hierenc_bias} shows that \textsc{HierEnc} with \model gains further improvements on generating QS hierarchies and full document summarization on \textsc{GovReport}. This points to promising future adoptions of \model by existing models that would benefit from encoding document structure. 

\begin{table}[t]
    \centering
    \small
    \setlength{\tabcolsep}{4pt}
    \begin{tabular}{lccccc}
    \toprule
        & \multicolumn{3}{c}{\textbf{QSGen-Hier}} & \multicolumn{2}{c}{\textbf{\textsc{GovReport}}} \\
        \textbf{Model} & \textbf{R-2} & \textbf{R-L} & \textbf{Hier F1} & \textbf{R-2} & \textbf{R-L} \\
        \midrule
        \textsc{HierEnc} & 16.32 & 38.06 & 11.77 & 28.83 & 56.99 \\
        w/ \model & +0.44 & +0.37 & +0.22 & +0.15 & +0.22 \\
        \bottomrule
    \end{tabular}
    \caption{Effects of applying \model to the extra section-level encoders of \textsc{HierEnc} on two tasks. 
    \model improves the performance of \textsc{HierEnc} on all metrics.
    }
    \label{tab:hierenc_bias}
\end{table}

%% file: 08_conclusion.tex
\section{Conclusion}

We present \model, which effectively and efficiently injects document structure information into abstractive summarization models via hierarchical learnable biases that adjust the attention score matrix. 
A new task, hierarchical question-summary generation, is then introduced for generating hierarchically organized question-summary pairs, to expose document structure and salient content to readers. 
We annotate a new dataset consisting of $6{,}153$ summary paragraphs with question-summary hierarchies to facilitate our study, and it can also be used for query-focused summarization and question generation. 
Experiments on hierarchical question-summary generation and full summary generation show that \model produces question-summary hierarchies of higher quality as measured by both automatic metrics and human judges, and achieves higher content coverage of summaries than competitive comparisons as reported by ROUGE.

%% file: acknowledgements.tex
\section*{Acknowledgements}
This work is supported in part by National Science Foundation through grant IIS-2046016, Oracle Cloud credits and related resources provided by the Oracle for Research program. 
We thank the anonymous reviewers for their valuable suggestions.

%% file: ethical_consideration.tex
\section*{Ethical Consideration}

\paragraph{Collection of \data and \textsc{WikiBioSum}.}

We comply with the terms of use and copyright policies of all data sources during the collection of \data and \textsc{WikiBioSum}.
Personal and other sensitive information is not collected to ensure the privacy of content creators.
Before annotating \data, we obtain consents from the annotators and inform them of their rights to temporarily suspend or quit the annotation process.
During annotation, annotators are fairly compensated ($\approx \$ 15$ per hour).

\paragraph{Limitations and Potential Risks of \model and \data.}

While our experiments focus on datasets consisting of formal long documents, we recognize that long documents could be written in informal languages where our model might not perform reasonably and could generate degraded or even incorrect outputs.
Despite recent advancement in improving summary factuality along with its evaluation~\cite{kryscinski-etal-2020-evaluating, goyal-durrett-2020-evaluating, scialom-etal-2021-questeval,cao-wang-2021-cliff}, the accuracy of existing factuality evaluation metrics has not been verified on long documents, which further increases the risk of incorrect outputs by our model.

As our \data is based on reports from the United States (US) Government, the topics covered by the dataset are mostly relevant to the national interest of US. Therefore, models trained on our dataset might not be suitable for producing structured summaries for documents published by other countries that focus on other topics.
Moreover, our \data might bias the model towards a pro-US perspective, which could produce outputs that are harmful to certain populations.

%% file: appendix_A_annotation.tex
\section{Details of \data}
\label{appendix:annotation}

\paragraph{Dataset Choice.}

We choose \textsc{GovReport} dataset~\cite{huang-etal-2021-efficient} for our annotation because it contains long documents (9409 tokens) and summaries (553 tokens) with key information spread throughout documents, which ensures the building of rich question-summary hierarchies. 
Moreover, the documents in \textsc{GovReport} are organized into multiple levels of sections, which justifies our decision to present salient document information with question-summary hierarchies.

\paragraph{Summary Paragraph Selection.}

Documents that are short or contain very few sections are less likely to yield rich QS hierarchies. 
To select high-quality paragraphs for annotation, we first consider using summary paragraphs associated with documents that have at least 3 sections. Moreover, the average number of paragraphs in each section should be at least 5. 
We then discard summaries that have less than 3 paragraphs. 
Among the paragraphs of the remaining summaries, we select those with at least 3 sentences and 70 words. 
To incorporate more abstractive summaries in the question-summary pairs, we further calculate the normalized density~\cite{grusky-etal-2018-newsroom} between each summary paragraph and its corresponding document, and then keep the paragraphs with a normalized density less than $0.15$. 
The selection process results in $25{,}063$ summary paragraphs which are then randomly sampled for annotation.

\paragraph{Annotation Process.}

We hire $11$ college students who are native English speakers as annotators. They are informed of the job opportunity through email lists that advertise on-campus jobs. They sign up for the annotation job by filling a Google Form containing a detailed job description and consent form. The employment process is handled through the school employment system.
Before annotating, they read the annotation instruction and examples with annotated question summary hierarchies. In each round of the annotation, each annotator is given 28--33 summary paragraphs, which takes about 2 hours to finish. We pay each annotator \$30 ($\approx$ \$15 per hour) for each round.
Appen\footnote{\url{https://appen.com}} is used for building the annotation interface and collecting annotations.
The annotation instruction is shown in Figure~\ref{fig:annotation_instr_p1}--\ref{fig:annotation_instr_p4}.

\paragraph{Section Alignment.}
We align each annotated summary paragraph with sections in the source document (\S~\ref{sec:hier_qs_gen}) in the following way. 
Three similarity scores are computed for each pair of summary sentence and document paragraph: (1) cosine similarity between the representations computed by Sentence BERT~\cite{reimers-2019-sentence-bert} for the summary sentence and the document paragraph; (2) the percentage of unique bigrams in the summary sentence that occur in the document paragraph; and (3) the percentage of unique named entities\footnote{We use SpaCy 3.0.3~\cite{spacy2} with \texttt{en\_core\_web\_sm} for named entity recognition.} that occur in the document paragraph.
The final similarity score is the sum of these three scores, with weights $0.4$, $1.0$, and $0.2$, respectively.
We tune the weights based on the manual alignment for $836$ summary paragraphs associated with $42$ report documents.
Finally, each summary sentence is mapped to the document paragraph with the highest similarity score.

\paragraph{Copyright Policy.}

Documents and summaries in \textsc{GovReport} dataset are published by Government Accountability Office (GAO)\footnote{\url{https://www.gao.gov/}} and Congressional Research Service (CRS)\footnote{\url{https://crsreports.congress.gov/}}.
The original publications are not protected by copyright law and \citet{huang-etal-2021-efficient} make \textsc{GovReport} publicly available.
We release the new annotations under the CC BY 4.0 license\footnote{\url{https://creativecommons.org/licenses/by/4.0/}}. Users of the data must also acknowledge GAO and CRS as the sources of the original publications.

%% file: appendix_B_data_detail.tex
\section{Details of \textsc{WikiBioSum}}
\label{appendix:dataset_detail}

\paragraph{Data Collection.}

To collect biographies from WikiProject Biography\footnote{\url{https://en.wikipedia.org/wiki/Wikipedia:WikiProject_Biography}}, we first use Scrapy\footnote{\url{https://scrapy.org}} to get the names of articles curated by the project.
We then extract article content with WikiExtractor\footnote{\url{https://github.com/attardi/wikiextractor}. We modify the original code so that full section structures can be preserved.} from the English Wikipedia dump\footnote{\url{https://dumps.wikimedia.org}} at 2021/08/01 using the article names.

\paragraph{Data Filtering.}

In addition to keeping biographies with at least two levels of section hierarchy, we discard biographies that have a quality class that lower than C.\footnote{Quality classes include FA, A, GA, B, C, Start, and Stub, sorted from best to worst.} The quality class of each biography is assessed by the members of WikiProject Biography.
To get rid of samples where summaries can be generated by reading the first half of the documents only, we check the occurrences of summary bigrams in the documents and keep the samples where the second half of the documents contain at least $9\%$ of new summary bigrams that do not occur in the first half.

\paragraph{Statistics.}

As reported in the main paper, the average lengths of the input and output are $3{,}478$ and $1{,}266$.
The average number of sections in the input is $11.65$, with an average depth of $2.22$ levels. 
Moreover, each document has $32.19$ paragraphs.

\paragraph{Copyright Policy.}

We follow the Wikipedia copyright policy\footnote{\url{https://en.wikipedia.org/wiki/Wikipedia:Copyrights}} to collect the \textsc{WikiBioSum} dataset. The \textsc{WikiBioSum} dataset will be released under the CC BY-SA 3.0 license\footnote{\url{https://creativecommons.org/licenses/by-sa/3.0/}}. 
Usage of the \textsc{WikiBioSum} dataset is limited by the copyright policy of Wikipedia.

%% file: appendix_C_human_eval.tex
\section{Details of Human Evaluation}
\label{appendix:human_eval}

We conduct human evaluation for question-summary hierarchies generated by five models.
Human evaluation instructions are shown in Figure~\ref{fig:human_eval_guideline}.
The annotators use an HTML interface (Figure~\ref{fig:human_eval_interface}).
Model names are not displayed, and their outputs in each group are randomly shuffled.
The interface displays all the annotations made by the same annotator, which helps human subjects achieve better annotation consistency across different model outputs. 
Finally, we report the average numbers of QS pairs per hierarchy for each model in Table~\ref{tab:human_eval_num_q}.

\begin{table}[t]
    \small
    \centering
    \begin{tabular}{lc}
    \toprule
        \textbf{model} & \textbf{Avg QS Pairs / Hier} \\
        \midrule
        \textsc{SecTok} & 5.29 \\
        \textsc{LvlSecTok} & 5.10 \\
        \textsc{HierEnc} & 5.29  \\
        \textsc{TokBias} & 4.95 \\
        \model-\textsc{enc} & 5.17 \\
        \bottomrule
    \end{tabular}
    \caption{Average numbers of QS pairs generated for each hierarchy by models in our human evaluation.}
    \label{tab:human_eval_num_q}
\end{table}

%% file: appendix_D_visualization.tex
\section{Additional Visualization}
\label{appendix:visualization}

\begin{figure}[t]
    \centering
    \includegraphics[width=0.48\textwidth]{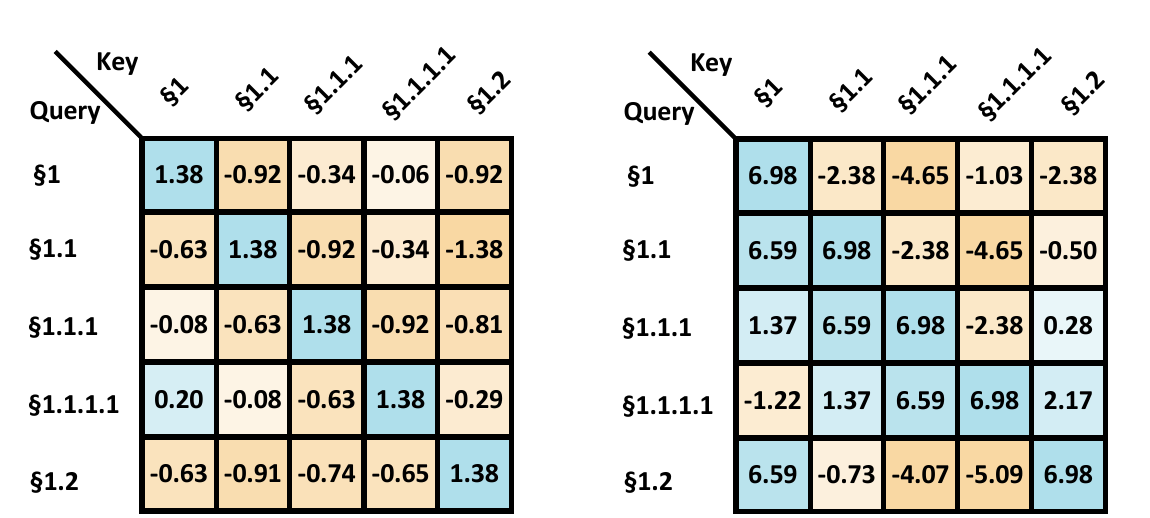}
    \caption{Visualization of hierarchical biases in \model-\textsc{enc} (left) and \model-\textsc{dec} (right) on \textsc{GovReport}. Positive and negative values are shaded in \hlc[positiveblue!49]{\textbf{blue}} and \hlc[negativeorange!49]{\textbf{orange}}. Displayed values are $100$X of actual values. 
    \model-\textsc{enc} biases towards recency, while \model-\textsc{dec} focuses on parent sections. 
    }
    \label{fig:bias_value_add}
\end{figure}

We show the biases learned by \model for full document summarization on \textsc{GovReport} in Figure~\ref{fig:bias_value_add}.
Behaviors of \model on \textsc{GovReport} are different from those observed on QSGen-Hier in \S\ref{sec:additional_study}. 
On \textsc{GovReport}, using \model on the encoder encourages each token to attend to other tokens within the same section, highlighting its focus on recency.
By contrast, \model on the decoder biases towards short-term contexts before a given token and strongly discourages attentions to long-range contexts.
It might be because that the generation of fluent and coherent summaries mainly depends on local and past contexts.

%% file: appendix_E_sample_output.tex
\section{Sample Output}

We show more outputs by \model-\textsc{Enc} on QSGen-Hier in Table~\ref{tab:sample_output}.

%% file: appendix_F_implementation.tex
\section{Details of Implementation}

We take the implementation of Longformer from Huggingface 4.8.1~\cite{wolf-etal-2020-transformers}, which is licensed under the Apache License 2.0\footnote{\url{https://www.apache.org/licenses/LICENSE-2.0}}. The model configuration and pre-trained weights of  \texttt{allenai/led-large-16384}\footnote{\url{https://huggingface.co/allenai/led-large-16384}} are used. For model training, we use Fairseq (commit \texttt{f34abcf2})~\cite{ott2019fairseq} that adopts MIT License\footnote{\url{https://opensource.org/licenses/MIT}}.
Both model training and decoding are performed on the A6000 GPU with 48GB memory and the A100 GPU with 40GB memory.

\paragraph{Training Settings.}

During training, we set the number of tokens in each batch to $10{,}240$ for QSGen-Hier, QSGen-ChildQ, and full summary generation on \textsc{WikiBioSum}.
On \textsc{GovReport}, each batch contains $16{,}384$ tokens.
As limited by the design of Longformer, the maximum output length for all tasks is set to $1{,}024$.
We use Adam~\cite{DBLP:journals/corr/KingmaB14} as the optimizer, with a maximum learning rate of $5 \times 10^{-5}$. The optimizer updates the model parameters every 8 batches.
We set the maximum numbers of update steps to $500$, $700$, $2{,}400$, and $5{,}000$ respectively for QSGen-Hier, QSGen-ChildQ, \textsc{WikiBioSum}, and \textsc{GovReport}.
Importantly, we adopt gradient checkpointing~\cite{chen2016training} to reduce the memory consumption of back propagation. 

\paragraph{Decoding Settings.}

A beam search with a beam size of $4$ is used for decoding. The maximum decoding length is $1{,}024$. We also disable the generation of repeated 5-grams.

\paragraph{Running Time.}

\model takes $2$, $2$, $5$, and $12$ hours for training on QSGen-Hier, QSGen-ChildQ, \textsc{WikiBioSum}, and \textsc{GovReport} with $4$ GPUs.
Decoding on QSGen-Hier and QSGen-ChildQ takes one hour. For decoding on \textsc{WikiBioSum}, and \textsc{GovReport}, it uses $3$ and $4$ hours.

\paragraph{Evaluation.}

We compute ROUGE scores~\cite{lin-2004-rouge} using the implementation by Google Research\footnote{\url{https://github.com/google-research/google-research/tree/master/rouge}}.
For BLEU scores, we use NLTK 3.5~\cite{bird2009natural}.

%% file: appendix_large_table_figure.tex
\begin{table*}[t]
    \centering
    \fontsize{10}{12}\selectfont
    \begin{tabular}{p{0.85\textwidth}}
        \toprule
        \textbf{Example 1} \\
        \midrule
        Q1: What incited the start of the FY2009 appropriation process? \\
        A1: On February 4, 2008, President Bush sent his FY2009 budget to Congress, which included a request for \$39 billion for the Department of Housing and Urban Development (HUD). \\
        \quad Q1.1: How did Congress respond to this request? \\
        \quad A1.1: On June 4, 2008, the Senate passed the FY2009 budget resolution conference agreement (H.Rept. 110-659) and the House passed it the following day. \\
        \quad Q1.2: What was the result of the FY2009 appropriations process? \\
        \quad A1.2: On March 11, 2009, a FY2009 omnibus appropriations bill was signed into law, funding HUD for the remainder of the fiscal year (P.L. 111-8). It provides a more than 10\% increase in regular, non-emergency appropriations over the FY2008 level. \\
        \quad \quad Q1.2.1: How did the omnibus appropriations bill affect HUD? \\
        \quad \quad A1.2.1: It provided nearly \$13.7 billion for HUD programs. \\
        \midrule
        \textbf{Example 2} \\
        \midrule
        Q1: To what extent is democracy promotion an element of U.S. foreign policy? \\
        A1: For decades U.S. policymakers have connected U.S. national security and other core interests with the spread of democracy around the world. Reflecting this, the promotion of democracy has been a longstanding and multifaceted element of U.S. foreign policy, and one often interrelated with U.S. efforts to promote human rights. \\
        \quad Q1.1: How has the promotion of democracy promotion been supported by Congress? \\
        \quad A1.1: Congress has often played an important role in supporting and institutionalizing U.S. democracy promotion by passing key legislation, appropriating funds for foreign assistance programs and other democracy promoting activities, and conducting oversight of aspects of U.S.-led foreign policy relevant to democracy promotion. \\
        \quad Q1.2: What is the current state of democracy promotion? \\
        \quad A1.2: Widespread concerns exist among analysts and policymakers over the current trajectory of democracy around theworld and multiple hearings in the 115th Congress reflected bipartisan concern over this issue. \\
        \quad \quad Q1.2.1: What are some of these concerns? \\
        \quad \quad A1.2.1: Frequently cited concerns include the rise of authoritarian populist and nationalist leaders, the potential negative influence on democracy from internationally assertive authoritarian states, questions over the enduring appeal of democracy as a political system, new tools nondemocratic governments are using to stifle potential democratizing forces, and others. \\
        \midrule
        \textbf{Example 3} \\
        \midrule
        Q1: How should GA strategies be approached? \\
        A1: GA security poses significant challenges for policymakers and security experts because GA is highly diverse, geographically dispersed, and relatively open compared to commercial airports servicing passenger airlines and other protected infrastructure such as nuclear reactors and chemical plants. \\
        Q2: What is the primary threat posed by GA aircraft? \\
        A2: The primary threat posed to GA aircraft is not so much to GA assets themselves, but rather, from terrorists seeking to exploit GA assets to attack critical infrastructure or high-profile targets. \\
        \quad Q2.1: What is a secondary threat to GA aircraft? \\
        \quad A2.1: A secondary threat is that terrorists may infiltrate or otherwise exploit GA to gain knowledge and/or access to the airspace system in the United States. \\
        \quad \quad Q2.1.1: What are some examples of this threat? \\
        \quad \quad A2.1.1: For example, some corporate aviation operators have expressed concern that aircraft carrying high-profile business leaders and executives, such as presidents of major U.S. corporations, could be targeted, particularly when operating overseas in areas where security concerns exist. \\
        \bottomrule
    \end{tabular}
    \caption{Example outputs by \model-\textsc{Enc} on QSGen-Hier. Indentation indicates the levels of question-summary pairs.}
    \label{tab:sample_output}
\end{table*}

\begin{figure*}[t]
    \centering
    \includegraphics[width=0.95\textwidth, page=1]{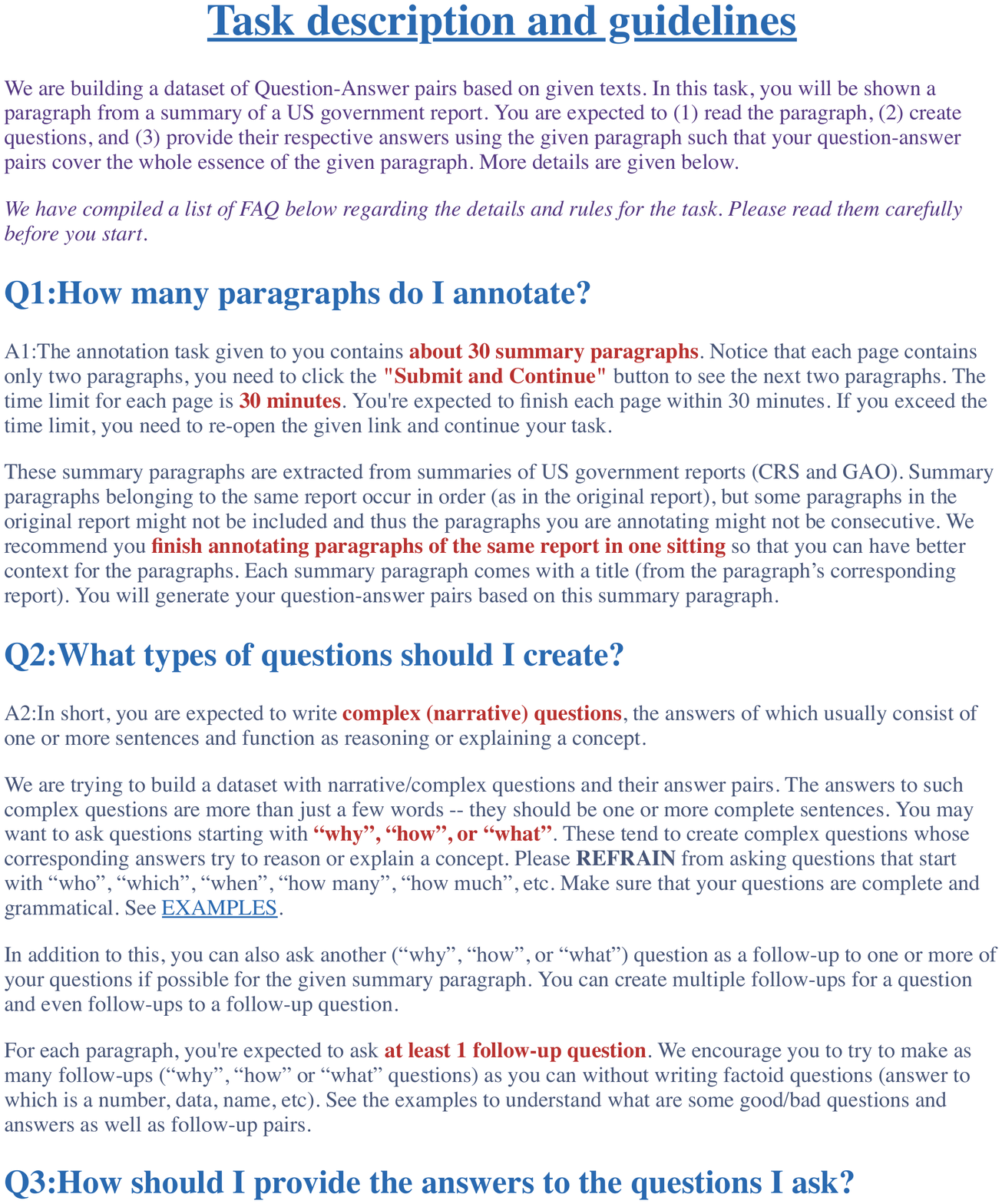}
    \caption{Question-summary hierarchy annotation instructions. (Page 1 / 4)}
    \label{fig:annotation_instr_p1}
\end{figure*}

\begin{figure*}[t]
    \centering
    \includegraphics[width=0.95\textwidth, page=2]{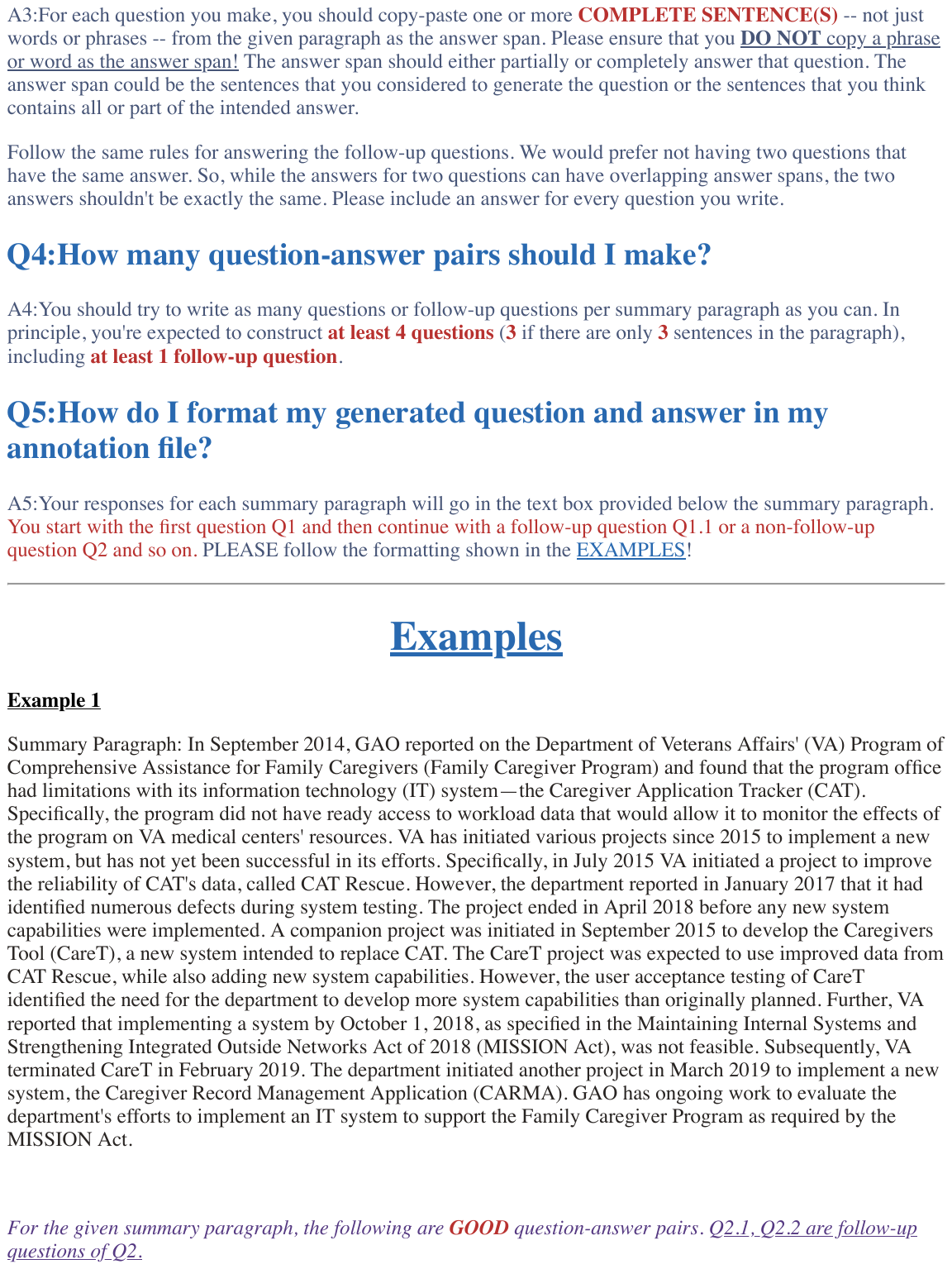}
    \caption{Question-summary hierarchy annotation instructions. (Page 2 / 4)}
    \label{fig:annotation_instr_p2}
\end{figure*}

\begin{figure*}[t]
    \centering
    \includegraphics[width=0.95\textwidth, page=3]{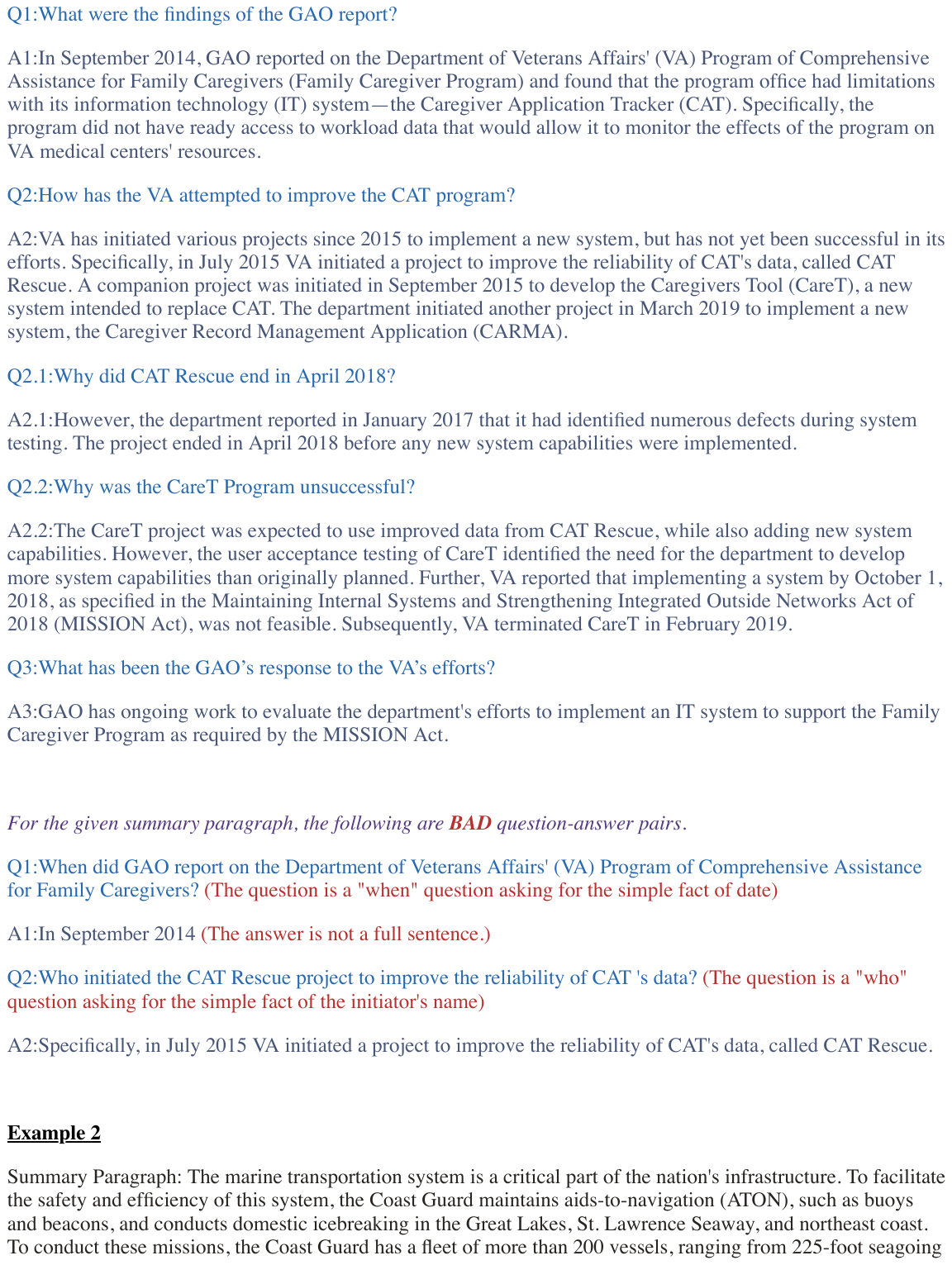}
    \caption{Question-summary hierarchy annotation instructions. (Page 3 / 4)}
    \label{fig:annotation_instr_p3}
\end{figure*}

\begin{figure*}[t]
    \centering
    \includegraphics[width=0.95\textwidth, page=4]{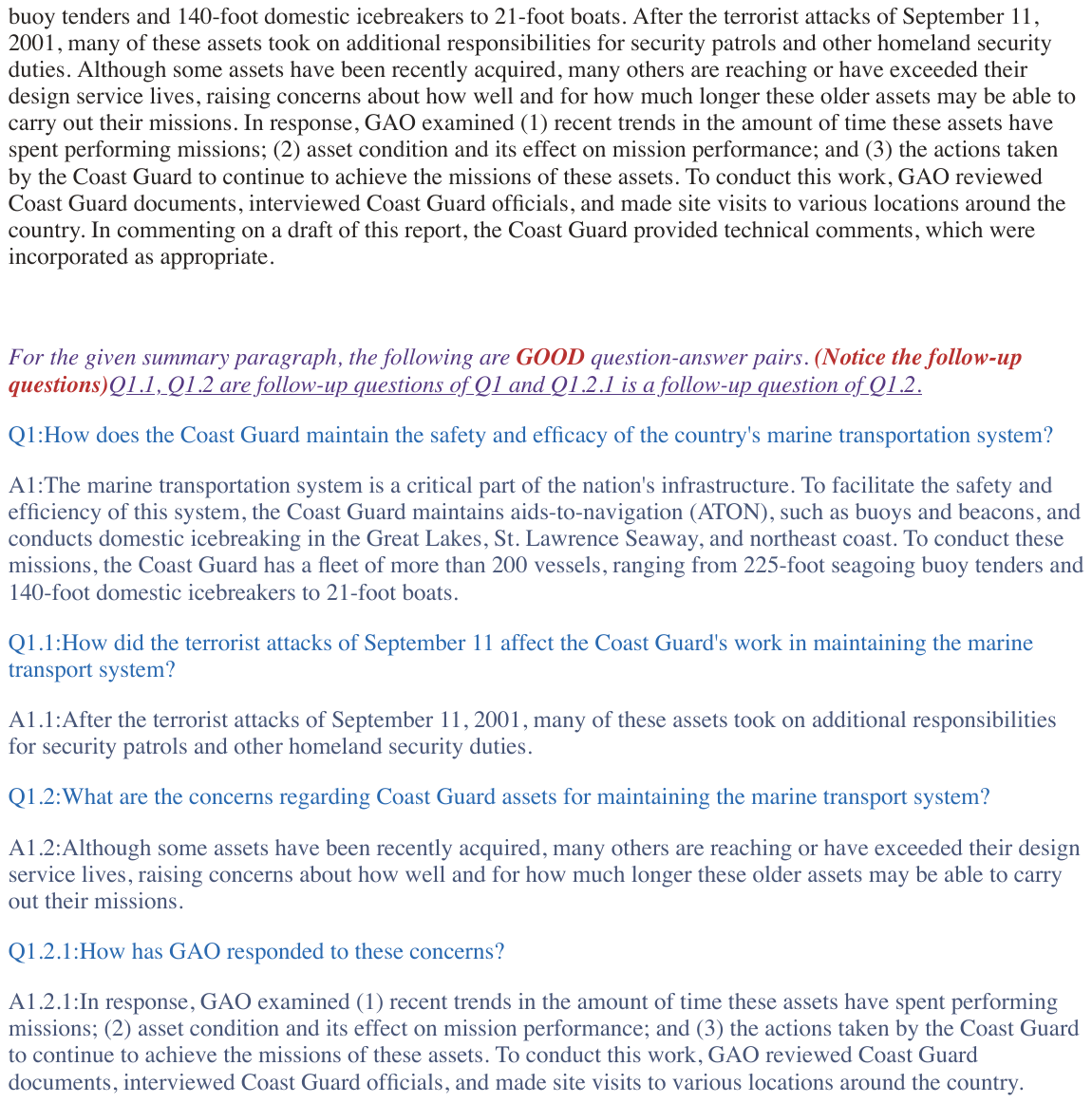}
    \caption{Question-summary hierarchy annotation instructions. (Page 4 / 4)}
    \label{fig:annotation_instr_p4}
\end{figure*}

\begin{figure*}[t]
    \centering
    \fontsize{10}{12}\selectfont
    \begin{tabular}{p{0.92\textwidth}}
    \toprule
        In this study, you will evaluate 50 sets of \textbf{question-summary (QS) hierarchies} produced by five systems. The hierarchy is presented by the IDs of questions and summaries (e.g., Q1 is the parent of Q1.1 and Q1.2). We also consider there is a dummy root to be the parent of the top-level questions (e.g., Q1, Q2). \\
        Please go through the hierarchy generated by each system in order. For each QS pair in the hierarchy, you need to \textbf{adjust it step by step} such that it has the most appropriate QS pair as its parent. Meanwhile, please also check if the question can be \textbf{answered} by its corresponding summary. The descriptions of how to make the adjustment and determine answerability are detailed as follows with an example. \\
        \midrule
        Example \\
        \midrule
        (DUMMY ROOT) \\
        
        Q1: What did state officials report about the effectiveness of identification verification procedures? \\
        A1: State officials interviewed by GAO report that identity verification procedures have been effective at combating certain kinds of fraud, but vulnerabilities remain. Officials in most of the 11 states GAO contacted reported a decline in the use of counterfeit identity documents, and officials in states using facial recognition said they detected a number of identity theft attempts. \\
        
        Q1.1: How can criminals use someone else's identity to get a license in another state? \\
        A1.1: However, criminals can still steal the identity of someone in one state and use it to get a license in another because states lack the capacity to consistently detect such cross-state fraud. \\
        
        Q1.1.1: What is one solution to this existing issue? \\
        A1.1.1: For example, one state officials told GAO a check against the problem driver database (Problem Driver Pointer System) will not detect a license in another state if it is not associated with any driving violation. \\
        
        Q2: ... \\
        A2: ... \\
        
        \smallskip
        \textbf{Step-by-step Adjustment:} In QS hierarchies, the children of a QS pair ask about follow-up information that could be specific descriptions or elaborations of the content in the QS pair. 
        For each QS pair, you need to first determine another QS pair (or the dummy root) as its parent such that they form the best follow-up relation. After identifying the most appropriate parent, adjustment of the QS pair is conducted step by step. In each step, you can attach the QS pair to its grandparent or sibling (i.e., the parent or child of its current parent). \\
        Please report the \textbf{number of steps} required to complete the adjustment. If no adjustment is needed, please report 0. \\
        For example, the most appropriate parent for Q1.1 is the DUMMY ROOT because it asks about a concrete flaw of the identification verification procedure while Q1 and A1 talk about the effectiveness of the procedure. These two questions are regarding the current status of the identification verification procedure and they should be at the same level.
        As there is an edge between Q1 and DUMMY ROOT, you only need \textbf{one} step to finish attaching Q1.1 to DUMMY ROOT. (Q1 $ \rightarrow $ DUMMY ROOT). \\
        \textbf{Note} that the parent-child relation remains unchanged for the children and descendants of an adjusted QS pair. For example, after attaching Q1.1 to DUMMY ROOT, attaching Q1.1.1 to Q2 only needs two steps as Q1.1 is already attached to DUMMY ROOT (Q1.1 $ \rightarrow $ DUMMY ROOT $ \rightarrow $ Q2).
        
        \smallskip
        \textbf{Answerability:} Whether the question can be answered by the associated summary. \\
        Please select ``True'' or ``False'' for each QS pair. \\
        For example, Q1.1.1 is not answerable because A1.1.1 does not mention any solution. Both Q1 and Q1.1 are answerable. \\

        \bottomrule
    \end{tabular}
    \caption{Human evaluation guidelines.}
    \label{fig:human_eval_guideline}
\end{figure*}

\begin{figure*}
    \centering
    \includegraphics[width=0.99\textwidth]{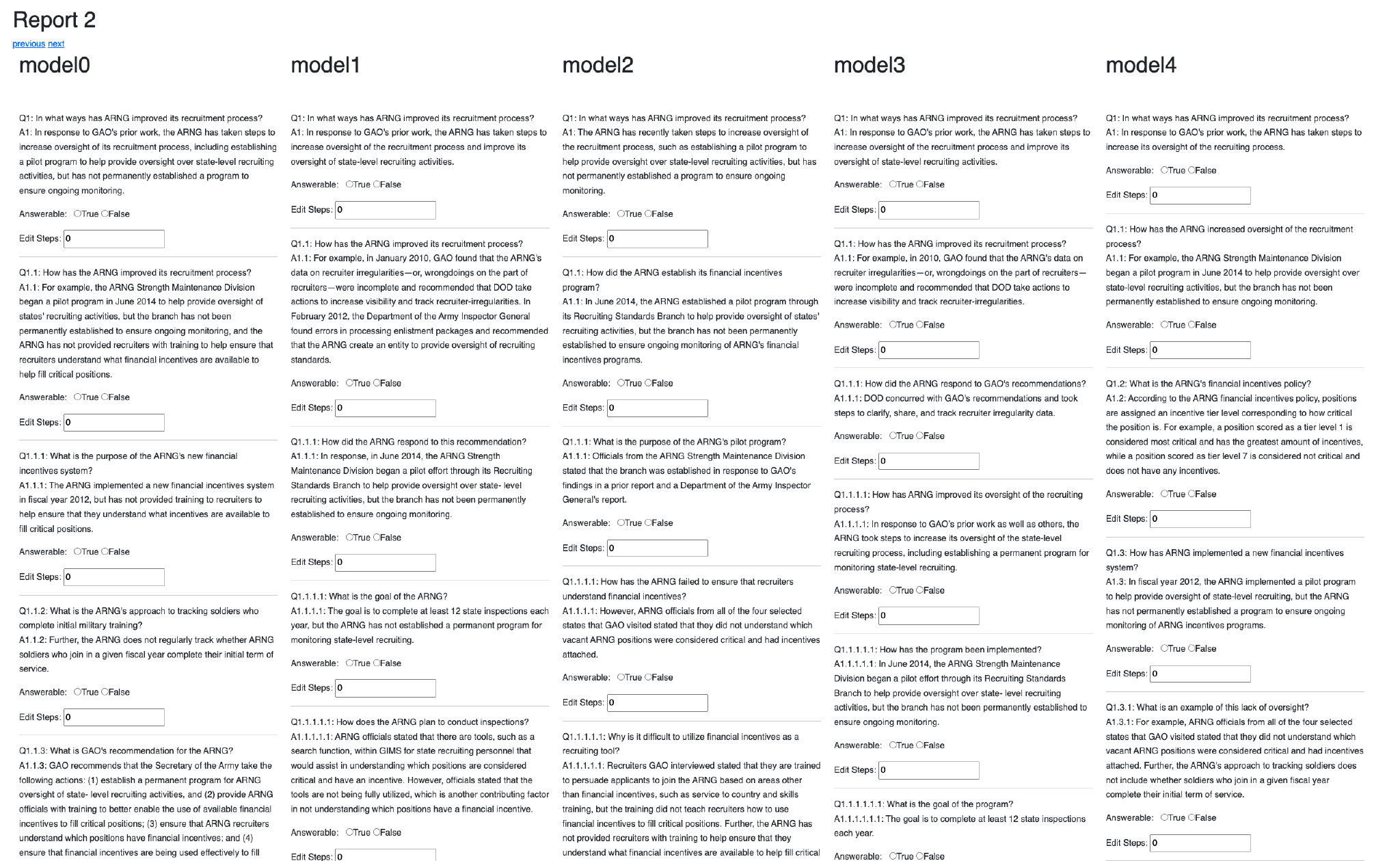}
    \caption{Screenshot of the human evaluation interface.}
    \label{fig:human_eval_interface}
\end{figure*}

%% file: acl_latex.bbl
\begin{thebibliography}{60}
\expandafter\ifx\csname natexlab\endcsname\relax\def\natexlab#1{#1}\fi

\bibitem[{Alberti et~al.(2019)Alberti, Andor, Pitler, Devlin, and
  Collins}]{alberti-etal-2019-synthetic}
Chris Alberti, Daniel Andor, Emily Pitler, Jacob Devlin, and Michael Collins.
  2019.
\newblock \href {https://doi.org/10.18653/v1/P19-1620} {Synthetic {QA} corpora
  generation with roundtrip consistency}.
\newblock In \emph{Proceedings of the 57th Annual Meeting of the Association
  for Computational Linguistics}, pages 6168--6173, Florence, Italy.
  Association for Computational Linguistics.

\bibitem[{Angelidis and Lapata(2018)}]{angelidis-lapata-2018-summarizing}
Stefanos Angelidis and Mirella Lapata. 2018.
\newblock \href {https://doi.org/10.18653/v1/D18-1403} {Summarizing opinions:
  Aspect extraction meets sentiment prediction and they are both weakly
  supervised}.
\newblock In \emph{Proceedings of the 2018 Conference on Empirical Methods in
  Natural Language Processing}, pages 3675--3686, Brussels, Belgium.
  Association for Computational Linguistics.

\bibitem[{Balachandran et~al.(2021)Balachandran, Pagnoni, Lee, Rajagopal,
  Carbonell, and Tsvetkov}]{balachandran-etal-2021-structsum}
Vidhisha Balachandran, Artidoro Pagnoni, Jay~Yoon Lee, Dheeraj Rajagopal, Jaime
  Carbonell, and Yulia Tsvetkov. 2021.
\newblock \href {https://doi.org/10.18653/v1/2021.eacl-main.220}
  {{S}truct{S}um: Summarization via structured representations}.
\newblock In \emph{Proceedings of the 16th Conference of the European Chapter
  of the Association for Computational Linguistics: Main Volume}, pages
  2575--2585, Online. Association for Computational Linguistics.

\bibitem[{Barzilay and Elhadad(1999)}]{barzilay1999using}
Regina Barzilay and Michael Elhadad. 1999.
\newblock Using lexical chains for text summarization.
\newblock \emph{Advances in automatic text summarization}, pages 111--121.

\bibitem[{Barzilay and Lee(2004)}]{barzilay-lee-2004-catching}
Regina Barzilay and Lillian Lee. 2004.
\newblock \href {https://aclanthology.org/N04-1015} {Catching the drift:
  Probabilistic content models, with applications to generation and
  summarization}.
\newblock In \emph{Proceedings of the Human Language Technology Conference of
  the North {A}merican Chapter of the Association for Computational
  Linguistics: {HLT}-{NAACL} 2004}, pages 113--120, Boston, Massachusetts, USA.
  Association for Computational Linguistics.

\bibitem[{Beltagy et~al.(2020)Beltagy, Peters, and
  Cohan}]{beltagy2020longformer}
Iz~Beltagy, Matthew~E. Peters, and Arman Cohan. 2020.
\newblock \href {http://arxiv.org/abs/2004.05150} {Longformer: The
  long-document transformer}.

\bibitem[{Bird et~al.(2009)Bird, Klein, and Loper}]{bird2009natural}
Steven Bird, Ewan Klein, and Edward Loper. 2009.
\newblock \emph{Natural language processing with Python: analyzing text with
  the natural language toolkit}.
\newblock " O'Reilly Media, Inc.".

\bibitem[{Cao and Wang(2021{\natexlab{a}})}]{cao-wang-2021-attention}
Shuyang Cao and Lu~Wang. 2021{\natexlab{a}}.
\newblock \href {https://doi.org/10.18653/v1/2021.naacl-main.397} {Attention
  head masking for inference time content selection in abstractive
  summarization}.
\newblock In \emph{Proceedings of the 2021 Conference of the North American
  Chapter of the Association for Computational Linguistics: Human Language
  Technologies}, pages 5008--5016, Online. Association for Computational
  Linguistics.

\bibitem[{Cao and Wang(2021{\natexlab{b}})}]{cao-wang-2021-cliff}
Shuyang Cao and Lu~Wang. 2021{\natexlab{b}}.
\newblock \href {https://doi.org/10.18653/v1/2021.emnlp-main.532} {{CLIFF}:
  Contrastive learning for improving faithfulness and factuality in abstractive
  summarization}.
\newblock In \emph{Proceedings of the 2021 Conference on Empirical Methods in
  Natural Language Processing}, pages 6633--6649, Online and Punta Cana,
  Dominican Republic. Association for Computational Linguistics.

\bibitem[{Celikyilmaz et~al.(2018)Celikyilmaz, Bosselut, He, and
  Choi}]{celikyilmaz-etal-2018-deep}
Asli Celikyilmaz, Antoine Bosselut, Xiaodong He, and Yejin Choi. 2018.
\newblock \href {https://doi.org/10.18653/v1/N18-1150} {Deep communicating
  agents for abstractive summarization}.
\newblock In \emph{Proceedings of the 2018 Conference of the North {A}merican
  Chapter of the Association for Computational Linguistics: Human Language
  Technologies, Volume 1 (Long Papers)}, pages 1662--1675, New Orleans,
  Louisiana. Association for Computational Linguistics.

\bibitem[{Chen et~al.(2016)Chen, Xu, Zhang, and Guestrin}]{chen2016training}
Tianqi Chen, Bing Xu, Chiyuan Zhang, and Carlos Guestrin. 2016.
\newblock \href {http://arxiv.org/abs/1604.06174} {Training deep nets with
  sublinear memory cost}.

\bibitem[{Choi et~al.(2018)Choi, He, Iyyer, Yatskar, Yih, Choi, Liang, and
  Zettlemoyer}]{choi-etal-2018-quac}
Eunsol Choi, He~He, Mohit Iyyer, Mark Yatskar, Wen-tau Yih, Yejin Choi, Percy
  Liang, and Luke Zettlemoyer. 2018.
\newblock \href {https://doi.org/10.18653/v1/D18-1241} {{Q}u{AC}: Question
  answering in context}.
\newblock In \emph{Proceedings of the 2018 Conference on Empirical Methods in
  Natural Language Processing}, pages 2174--2184, Brussels, Belgium.
  Association for Computational Linguistics.

\bibitem[{Cohan et~al.(2018)Cohan, Dernoncourt, Kim, Bui, Kim, Chang, and
  Goharian}]{cohan-etal-2018-discourse}
Arman Cohan, Franck Dernoncourt, Doo~Soon Kim, Trung Bui, Seokhwan Kim, Walter
  Chang, and Nazli Goharian. 2018.
\newblock \href {https://doi.org/10.18653/v1/N18-2097} {A discourse-aware
  attention model for abstractive summarization of long documents}.
\newblock In \emph{Proceedings of the 2018 Conference of the North {A}merican
  Chapter of the Association for Computational Linguistics: Human Language
  Technologies, Volume 2 (Short Papers)}, pages 615--621, New Orleans,
  Louisiana. Association for Computational Linguistics.

\bibitem[{Cui and Hu(2021)}]{cui-hu-2021-sliding}
Peng Cui and Le~Hu. 2021.
\newblock \href {https://doi.org/10.18653/v1/2021.naacl-main.470} {Sliding
  selector network with dynamic memory for extractive summarization of long
  documents}.
\newblock In \emph{Proceedings of the 2021 Conference of the North American
  Chapter of the Association for Computational Linguistics: Human Language
  Technologies}, pages 5881--5891, Online. Association for Computational
  Linguistics.

\bibitem[{Daum{\'e}~III and Marcu(2006)}]{daume2006bayesian}
Hal Daum{\'e}~III and Daniel Marcu. 2006.
\newblock Bayesian query-focused summarization.
\newblock In \emph{Proceedings of the 21st International Conference on
  Computational Linguistics and the 44th annual meeting of the Association for
  Computational Linguistics}, pages 305--312.

\bibitem[{Du and Cardie(2018)}]{du-cardie-2018-harvesting}
Xinya Du and Claire Cardie. 2018.
\newblock \href {https://doi.org/10.18653/v1/P18-1177} {Harvesting
  paragraph-level question-answer pairs from {W}ikipedia}.
\newblock In \emph{Proceedings of the 56th Annual Meeting of the Association
  for Computational Linguistics (Volume 1: Long Papers)}, pages 1907--1917,
  Melbourne, Australia. Association for Computational Linguistics.

\bibitem[{Durrett et~al.(2016)Durrett, Berg-Kirkpatrick, and
  Klein}]{durrett-etal-2016-learning}
Greg Durrett, Taylor Berg-Kirkpatrick, and Dan Klein. 2016.
\newblock \href {https://doi.org/10.18653/v1/P16-1188} {Learning-based
  single-document summarization with compression and anaphoricity constraints}.
\newblock In \emph{Proceedings of the 54th Annual Meeting of the Association
  for Computational Linguistics (Volume 1: Long Papers)}, pages 1998--2008,
  Berlin, Germany. Association for Computational Linguistics.

\bibitem[{Garg et~al.(2019)Garg, Peitz, Nallasamy, and
  Paulik}]{garg-etal-2019-jointly}
Sarthak Garg, Stephan Peitz, Udhyakumar Nallasamy, and Matthias Paulik. 2019.
\newblock \href {https://doi.org/10.18653/v1/D19-1453} {Jointly learning to
  align and translate with transformer models}.
\newblock In \emph{Proceedings of the 2019 Conference on Empirical Methods in
  Natural Language Processing and the 9th International Joint Conference on
  Natural Language Processing (EMNLP-IJCNLP)}, pages 4453--4462, Hong Kong,
  China. Association for Computational Linguistics.

\bibitem[{Goyal and Durrett(2020)}]{goyal-durrett-2020-evaluating}
Tanya Goyal and Greg Durrett. 2020.
\newblock \href {https://doi.org/10.18653/v1/2020.findings-emnlp.322}
  {Evaluating factuality in generation with dependency-level entailment}.
\newblock In \emph{Findings of the Association for Computational Linguistics:
  EMNLP 2020}, pages 3592--3603, Online. Association for Computational
  Linguistics.

\bibitem[{Grusky et~al.(2018)Grusky, Naaman, and
  Artzi}]{grusky-etal-2018-newsroom}
Max Grusky, Mor Naaman, and Yoav Artzi. 2018.
\newblock \href {https://doi.org/10.18653/v1/N18-1065} {{N}ewsroom: A dataset
  of 1.3 million summaries with diverse extractive strategies}.
\newblock In \emph{Proceedings of the 2018 Conference of the North {A}merican
  Chapter of the Association for Computational Linguistics: Human Language
  Technologies, Volume 1 (Long Papers)}, pages 708--719, New Orleans,
  Louisiana. Association for Computational Linguistics.

\bibitem[{Guthrie et~al.(1991)Guthrie, Britten, and Barker}]{10.2307/747765}
John~T. Guthrie, Tracy Britten, and K.~Georgene Barker. 1991.
\newblock \href {http://www.jstor.org/stable/747765} {Roles of document
  structure, cognitive strategy, and awareness in searching for information}.
\newblock \emph{Reading Research Quarterly}, 26(3):300--324.

\bibitem[{Hintikka(1981)}]{hintikka1981logic}
Jaakko Hintikka. 1981.
\newblock The logic of information-seeking dialogues: A model.

\bibitem[{Hirao et~al.(2013)Hirao, Yoshida, Nishino, Yasuda, and
  Nagata}]{hirao-etal-2013-single}
Tsutomu Hirao, Yasuhisa Yoshida, Masaaki Nishino, Norihito Yasuda, and Masaaki
  Nagata. 2013.
\newblock \href {https://aclanthology.org/D13-1158} {Single-document
  summarization as a tree knapsack problem}.
\newblock In \emph{Proceedings of the 2013 Conference on Empirical Methods in
  Natural Language Processing}, pages 1515--1520, Seattle, Washington, USA.
  Association for Computational Linguistics.

\bibitem[{Honnibal and Montani(2017)}]{spacy2}
Matthew Honnibal and Ines Montani. 2017.
\newblock {spaCy 2}: Natural language understanding with {B}loom embeddings,
  convolutional neural networks and incremental parsing.
\newblock To appear.

\bibitem[{Huang et~al.(2021)Huang, Cao, Parulian, Ji, and
  Wang}]{huang-etal-2021-efficient}
Luyang Huang, Shuyang Cao, Nikolaus Parulian, Heng Ji, and Lu~Wang. 2021.
\newblock \href {https://doi.org/10.18653/v1/2021.naacl-main.112} {Efficient
  attentions for long document summarization}.
\newblock In \emph{Proceedings of the 2021 Conference of the North American
  Chapter of the Association for Computational Linguistics: Human Language
  Technologies}, pages 1419--1436, Online. Association for Computational
  Linguistics.

\bibitem[{Jonassen(1988)}]{10.2307/44426154}
David~H. Jonassen. 1988.
\newblock \href {http://www.jstor.org/stable/44426154} {Designing structured
  hypertext and structuring access to hypertext}.
\newblock \emph{Educational Technology}, 28(11):13--16.

\bibitem[{Kingma and Ba(2015)}]{DBLP:journals/corr/KingmaB14}
Diederik~P. Kingma and Jimmy Ba. 2015.
\newblock \href {http://arxiv.org/abs/1412.6980} {Adam: {A} method for
  stochastic optimization}.
\newblock In \emph{3rd International Conference on Learning Representations,
  {ICLR} 2015, San Diego, CA, USA, May 7-9, 2015, Conference Track
  Proceedings}.

\bibitem[{Krishna and Iyyer(2019)}]{krishna-iyyer-2019-generating}
Kalpesh Krishna and Mohit Iyyer. 2019.
\newblock \href {https://doi.org/10.18653/v1/P19-1224} {Generating
  question-answer hierarchies}.
\newblock In \emph{Proceedings of the 57th Annual Meeting of the Association
  for Computational Linguistics}, pages 2321--2334, Florence, Italy.
  Association for Computational Linguistics.

\bibitem[{Kryscinski et~al.(2020)Kryscinski, McCann, Xiong, and
  Socher}]{kryscinski-etal-2020-evaluating}
Wojciech Kryscinski, Bryan McCann, Caiming Xiong, and Richard Socher. 2020.
\newblock \href {https://doi.org/10.18653/v1/2020.emnlp-main.750} {Evaluating
  the factual consistency of abstractive text summarization}.
\newblock In \emph{Proceedings of the 2020 Conference on Empirical Methods in
  Natural Language Processing (EMNLP)}, pages 9332--9346, Online. Association
  for Computational Linguistics.

\bibitem[{Kryściński et~al.(2021)Kryściński, Rajani, Agarwal, Xiong, and
  Radev}]{kryscinski2021booksum}
Wojciech Kryściński, Nazneen Rajani, Divyansh Agarwal, Caiming Xiong, and
  Dragomir Radev. 2021.
\newblock \href {http://arxiv.org/abs/2105.08209} {Booksum: A collection of
  datasets for long-form narrative summarization}.

\bibitem[{Lewis et~al.(2020)Lewis, Liu, Goyal, Ghazvininejad, Mohamed, Levy,
  Stoyanov, and Zettlemoyer}]{lewis-etal-2020-bart}
Mike Lewis, Yinhan Liu, Naman Goyal, Marjan Ghazvininejad, Abdelrahman Mohamed,
  Omer Levy, Veselin Stoyanov, and Luke Zettlemoyer. 2020.
\newblock \href {https://doi.org/10.18653/v1/2020.acl-main.703} {{BART}:
  Denoising sequence-to-sequence pre-training for natural language generation,
  translation, and comprehension}.
\newblock In \emph{Proceedings of the 58th Annual Meeting of the Association
  for Computational Linguistics}, pages 7871--7880, Online. Association for
  Computational Linguistics.

\bibitem[{Lin(2004)}]{lin-2004-rouge}
Chin-Yew Lin. 2004.
\newblock \href {https://aclanthology.org/W04-1013} {{ROUGE}: A package for
  automatic evaluation of summaries}.
\newblock In \emph{Text Summarization Branches Out}, pages 74--81, Barcelona,
  Spain. Association for Computational Linguistics.

\bibitem[{Liu et~al.(2020)Liu, Wei, Niu, Chen, and
  He}]{10.1145/3366423.3380270}
Bang Liu, Haojie Wei, Di~Niu, Haolan Chen, and Yancheng He. 2020.
\newblock \href {https://doi.org/10.1145/3366423.3380270} {\emph{Asking
  Questions the Human Way: Scalable Question-Answer Generation from Text
  Corpus}}, page 2032–2043. Association for Computing Machinery, New York,
  NY, USA.

\bibitem[{Liu and Lapata(2019)}]{liu-lapata-2019-hierarchical}
Yang Liu and Mirella Lapata. 2019.
\newblock \href {https://doi.org/10.18653/v1/P19-1500} {Hierarchical
  transformers for multi-document summarization}.
\newblock In \emph{Proceedings of the 57th Annual Meeting of the Association
  for Computational Linguistics}, pages 5070--5081, Florence, Italy.
  Association for Computational Linguistics.

\bibitem[{Liu et~al.(2019)Liu, Titov, and Lapata}]{liu-etal-2019-single}
Yang Liu, Ivan Titov, and Mirella Lapata. 2019.
\newblock \href {https://doi.org/10.18653/v1/N19-1173} {Single document
  summarization as tree induction}.
\newblock In \emph{Proceedings of the 2019 Conference of the North {A}merican
  Chapter of the Association for Computational Linguistics: Human Language
  Technologies, Volume 1 (Long and Short Papers)}, pages 1745--1755,
  Minneapolis, Minnesota. Association for Computational Linguistics.

\bibitem[{Marcu(1997)}]{marcu-1997-discourse}
Daniel Marcu. 1997.
\newblock \href {https://aclanthology.org/W97-0713} {From discourse structures
  to text summaries}.
\newblock In \emph{Intelligent Scalable Text Summarization}.

\bibitem[{McKeown et~al.(2009)McKeown, Beck, and Blake}]{10.2307/25655454}
Margaret~G. McKeown, Isabel~L. Beck, and Ronette~G.K. Blake. 2009.
\newblock \href {http://www.jstor.org/stable/25655454} {Rethinking reading
  comprehension instruction: A comparison of instruction for strategies and
  content approaches}.
\newblock \emph{Reading Research Quarterly}, 44(3):218--253.

\bibitem[{Meng et~al.(2021)Meng, Thaker, Zhang, Dong, Yuan, Wang, and
  He}]{meng-etal-2021-bringing}
Rui Meng, Khushboo Thaker, Lei Zhang, Yue Dong, Xingdi Yuan, Tong Wang, and
  Daqing He. 2021.
\newblock \href {https://doi.org/10.18653/v1/2021.acl-short.137} {Bringing
  structure into summaries: a faceted summarization dataset for long scientific
  documents}.
\newblock In \emph{Proceedings of the 59th Annual Meeting of the Association
  for Computational Linguistics and the 11th International Joint Conference on
  Natural Language Processing (Volume 2: Short Papers)}, pages 1080--1089,
  Online. Association for Computational Linguistics.

\bibitem[{Meyer et~al.(1980)Meyer, Brandt, and Bluth}]{10.2307/747349}
Bonnie J.~F. Meyer, David~M. Brandt, and George~J. Bluth. 1980.
\newblock \href {http://www.jstor.org/stable/747349} {Use of top-level
  structure in text: Key for reading comprehension of ninth-grade students}.
\newblock \emph{Reading Research Quarterly}, 16(1):72--103.

\bibitem[{Ott et~al.(2019)Ott, Edunov, Baevski, Fan, Gross, Ng, Grangier, and
  Auli}]{ott2019fairseq}
Myle Ott, Sergey Edunov, Alexei Baevski, Angela Fan, Sam Gross, Nathan Ng,
  David Grangier, and Michael Auli. 2019.
\newblock fairseq: A fast, extensible toolkit for sequence modeling.
\newblock In \emph{Proceedings of NAACL-HLT 2019: Demonstrations}.

\bibitem[{Papineni et~al.(2002)Papineni, Roukos, Ward, and
  Zhu}]{papineni-etal-2002-bleu}
Kishore Papineni, Salim Roukos, Todd Ward, and Wei-Jing Zhu. 2002.
\newblock \href {https://doi.org/10.3115/1073083.1073135} {{B}leu: a method for
  automatic evaluation of machine translation}.
\newblock In \emph{Proceedings of the 40th Annual Meeting of the Association
  for Computational Linguistics}, pages 311--318, Philadelphia, Pennsylvania,
  USA. Association for Computational Linguistics.

\bibitem[{Qu et~al.(2021)Qu, Jia, and Wu}]{qu-etal-2021-asking}
Fanyi Qu, Xin Jia, and Yunfang Wu. 2021.
\newblock \href {https://doi.org/10.18653/v1/2021.emnlp-main.202} {Asking
  questions like educational experts: {A}utomatically generating
  question-answer pairs on real-world examination data}.
\newblock In \emph{Proceedings of the 2021 Conference on Empirical Methods in
  Natural Language Processing}, pages 2583--2593, Online and Punta Cana,
  Dominican Republic. Association for Computational Linguistics.

\bibitem[{Raffel et~al.(2020)Raffel, Shazeer, Roberts, Lee, Narang, Matena,
  Zhou, Li, and Liu}]{JMLR:v21:20-074}
Colin Raffel, Noam Shazeer, Adam Roberts, Katherine Lee, Sharan Narang, Michael
  Matena, Yanqi Zhou, Wei Li, and Peter~J. Liu. 2020.
\newblock \href {http://jmlr.org/papers/v21/20-074.html} {Exploring the limits
  of transfer learning with a unified text-to-text transformer}.
\newblock \emph{Journal of Machine Learning Research}, 21(140):1--67.

\bibitem[{Rajpurkar et~al.(2016)Rajpurkar, Zhang, Lopyrev, and
  Liang}]{rajpurkar-etal-2016-squad}
Pranav Rajpurkar, Jian Zhang, Konstantin Lopyrev, and Percy Liang. 2016.
\newblock \href {https://doi.org/10.18653/v1/D16-1264} {{SQ}u{AD}: 100,000+
  questions for machine comprehension of text}.
\newblock In \emph{Proceedings of the 2016 Conference on Empirical Methods in
  Natural Language Processing}, pages 2383--2392, Austin, Texas. Association
  for Computational Linguistics.

\bibitem[{Reimers and Gurevych(2019)}]{reimers-2019-sentence-bert}
Nils Reimers and Iryna Gurevych. 2019.
\newblock \href {https://arxiv.org/abs/1908.10084} {Sentence-bert: Sentence
  embeddings using siamese bert-networks}.
\newblock In \emph{Proceedings of the 2019 Conference on Empirical Methods in
  Natural Language Processing}. Association for Computational Linguistics.

\bibitem[{Rohde et~al.(2021)Rohde, Wu, and Liu}]{rohde2021hierarchical}
Tobias Rohde, Xiaoxia Wu, and Yinhan Liu. 2021.
\newblock \href {http://arxiv.org/abs/2104.07545} {Hierarchical learning for
  generation with long source sequences}.

\bibitem[{Sachan and Xing(2018)}]{sachan-xing-2018-self}
Mrinmaya Sachan and Eric Xing. 2018.
\newblock \href {https://doi.org/10.18653/v1/N18-1058} {Self-training for
  jointly learning to ask and answer questions}.
\newblock In \emph{Proceedings of the 2018 Conference of the North {A}merican
  Chapter of the Association for Computational Linguistics: Human Language
  Technologies, Volume 1 (Long Papers)}, pages 629--640, New Orleans,
  Louisiana. Association for Computational Linguistics.

\bibitem[{Scialom et~al.(2021)Scialom, Dray, Lamprier, Piwowarski, Staiano,
  Wang, and Gallinari}]{scialom-etal-2021-questeval}
Thomas Scialom, Paul-Alexis Dray, Sylvain Lamprier, Benjamin Piwowarski, Jacopo
  Staiano, Alex Wang, and Patrick Gallinari. 2021.
\newblock \href {https://doi.org/10.18653/v1/2021.emnlp-main.529}
  {{Q}uest{E}val: Summarization asks for fact-based evaluation}.
\newblock In \emph{Proceedings of the 2021 Conference on Empirical Methods in
  Natural Language Processing}, pages 6594--6604, Online and Punta Cana,
  Dominican Republic. Association for Computational Linguistics.

\bibitem[{Shahaf et~al.(2012)Shahaf, Guestrin, and Horvitz}]{shahaf2012trains}
Dafna Shahaf, Carlos Guestrin, and Eric Horvitz. 2012.
\newblock Trains of thought: Generating information maps.
\newblock In \emph{Proceedings of the 21st international conference on World
  Wide Web}, pages 899--908.

\bibitem[{Shavelson(1974)}]{https://doi.org/10.1002/tea.3660110307}
Richard~J. Shavelson. 1974.
\newblock \href {https://doi.org/https://doi.org/10.1002/tea.3660110307}
  {Methods for examining representations of a subject-matter structure in a
  student's memory}.
\newblock \emph{Journal of Research in Science Teaching}, 11(3):231--249.

\bibitem[{Stede and Schlangen(2004)}]{stede2004information}
Manfred Stede and David Schlangen. 2004.
\newblock Information-seeking chat: Dialogues driven by topic-structure.
\newblock In \emph{Proceedings of Catalog (the 8th workshop on the semantics
  and pragmatics of dialogue; SemDial04)}. Citeseer.

\bibitem[{Taylor and Beach(1984)}]{10.2307/747358}
Barbara~M. Taylor and Richard~W. Beach. 1984.
\newblock \href {http://www.jstor.org/stable/747358} {The effects of text
  structure instruction on middle-grade students' comprehension and production
  of expository text}.
\newblock \emph{Reading Research Quarterly}, 19(2):134--146.

\bibitem[{Wang et~al.(2015)Wang, Cardie, and
  Marchetti}]{wang-etal-2015-socially}
Lu~Wang, Claire Cardie, and Galen Marchetti. 2015.
\newblock \href {https://doi.org/10.3115/v1/N15-1112} {Socially-informed
  timeline generation for complex events}.
\newblock In \emph{Proceedings of the 2015 Conference of the North {A}merican
  Chapter of the Association for Computational Linguistics: Human Language
  Technologies}, pages 1055--1065, Denver, Colorado. Association for
  Computational Linguistics.

\bibitem[{Wolf et~al.(2020)Wolf, Debut, Sanh, Chaumond, Delangue, Moi, Cistac,
  Rault, Louf, Funtowicz, Davison, Shleifer, von Platen, Ma, Jernite, Plu, Xu,
  Scao, Gugger, Drame, Lhoest, and Rush}]{wolf-etal-2020-transformers}
Thomas Wolf, Lysandre Debut, Victor Sanh, Julien Chaumond, Clement Delangue,
  Anthony Moi, Pierric Cistac, Tim Rault, Rémi Louf, Morgan Funtowicz, Joe
  Davison, Sam Shleifer, Patrick von Platen, Clara Ma, Yacine Jernite, Julien
  Plu, Canwen Xu, Teven~Le Scao, Sylvain Gugger, Mariama Drame, Quentin Lhoest,
  and Alexander~M. Rush. 2020.
\newblock \href {https://www.aclweb.org/anthology/2020.emnlp-demos.6}
  {Transformers: State-of-the-art natural language processing}.
\newblock In \emph{Proceedings of the 2020 Conference on Empirical Methods in
  Natural Language Processing: System Demonstrations}, pages 38--45, Online.
  Association for Computational Linguistics.

\bibitem[{Xiao and Carenini(2019)}]{xiao-carenini-2019-extractive}
Wen Xiao and Giuseppe Carenini. 2019.
\newblock \href {https://doi.org/10.18653/v1/D19-1298} {Extractive
  summarization of long documents by combining global and local context}.
\newblock In \emph{Proceedings of the 2019 Conference on Empirical Methods in
  Natural Language Processing and the 9th International Joint Conference on
  Natural Language Processing (EMNLP-IJCNLP)}, pages 3011--3021, Hong Kong,
  China. Association for Computational Linguistics.

\bibitem[{Xu et~al.(2020)Xu, Gan, Cheng, and Liu}]{xu-etal-2020-discourse}
Jiacheng Xu, Zhe Gan, Yu~Cheng, and Jingjing Liu. 2020.
\newblock \href {https://doi.org/10.18653/v1/2020.acl-main.451}
  {Discourse-aware neural extractive text summarization}.
\newblock In \emph{Proceedings of the 58th Annual Meeting of the Association
  for Computational Linguistics}, pages 5021--5031, Online. Association for
  Computational Linguistics.

\bibitem[{Zaheer et~al.(2020)Zaheer, Guruganesh, Dubey, Ainslie, Alberti,
  Ontanon, Pham, Ravula, Wang, Yang et~al.}]{zaheer2020bigbird}
Manzil Zaheer, Guru Guruganesh, Kumar~Avinava Dubey, Joshua Ainslie, Chris
  Alberti, Santiago Ontanon, Philip Pham, Anirudh Ravula, Qifan Wang, Li~Yang,
  et~al. 2020.
\newblock Big bird: Transformers for longer sequences.
\newblock \emph{Advances in Neural Information Processing Systems}, 33.

\bibitem[{Zeman et~al.(2017)Zeman, Popel, Straka, Haji{\v{c}}, Nivre, Ginter,
  Luotolahti, Pyysalo, Petrov, Potthast, Tyers, Badmaeva, Gokirmak, Nedoluzhko,
  Cinkov{\'a}, Haji{\v{c}}~jr., Hlav{\'a}{\v{c}}ov{\'a}, Kettnerov{\'a},
  Ure{\v{s}}ov{\'a}, Kanerva, Ojala, Missil{\"a}, Manning, Schuster, Reddy,
  Taji, Habash, Leung, de~Marneffe, Sanguinetti, Simi, Kanayama, de~Paiva,
  Droganova, Mart{\'\i}nez~Alonso, {\c{C}}{\"o}ltekin, Sulubacak, Uszkoreit,
  Macketanz, Burchardt, Harris, Marheinecke, Rehm, Kayadelen, Attia, Elkahky,
  Yu, Pitler, Lertpradit, Mandl, Kirchner, Alcalde, Strnadov{\'a}, Banerjee,
  Manurung, Stella, Shimada, Kwak, Mendon{\c{c}}a, Lando, Nitisaroj, and
  Li}]{zeman-etal-2017-conll}
Daniel Zeman, Martin Popel, Milan Straka, Jan Haji{\v{c}}, Joakim Nivre, Filip
  Ginter, Juhani Luotolahti, Sampo Pyysalo, Slav Petrov, Martin Potthast,
  Francis Tyers, Elena Badmaeva, Memduh Gokirmak, Anna Nedoluzhko, Silvie
  Cinkov{\'a}, Jan Haji{\v{c}}~jr., Jaroslava Hlav{\'a}{\v{c}}ov{\'a},
  V{\'a}clava Kettnerov{\'a}, Zde{\v{n}}ka Ure{\v{s}}ov{\'a}, Jenna Kanerva,
  Stina Ojala, Anna Missil{\"a}, Christopher~D. Manning, Sebastian Schuster,
  Siva Reddy, Dima Taji, Nizar Habash, Herman Leung, Marie-Catherine
  de~Marneffe, Manuela Sanguinetti, Maria Simi, Hiroshi Kanayama, Valeria
  de~Paiva, Kira Droganova, H{\'e}ctor Mart{\'\i}nez~Alonso,
  {\c{C}}a{\u{g}}r{\i} {\c{C}}{\"o}ltekin, Umut Sulubacak, Hans Uszkoreit,
  Vivien Macketanz, Aljoscha Burchardt, Kim Harris, Katrin Marheinecke, Georg
  Rehm, Tolga Kayadelen, Mohammed Attia, Ali Elkahky, Zhuoran Yu, Emily Pitler,
  Saran Lertpradit, Michael Mandl, Jesse Kirchner, Hector~Fernandez Alcalde,
  Jana Strnadov{\'a}, Esha Banerjee, Ruli Manurung, Antonio Stella, Atsuko
  Shimada, Sookyoung Kwak, Gustavo Mendon{\c{c}}a, Tatiana Lando, Rattima
  Nitisaroj, and Josie Li. 2017.
\newblock \href {https://doi.org/10.18653/v1/K17-3001} {{C}o{NLL} 2017 shared
  task: Multilingual parsing from raw text to {U}niversal {D}ependencies}.
\newblock In \emph{Proceedings of the {C}o{NLL} 2017 Shared Task: Multilingual
  Parsing from Raw Text to Universal Dependencies}, pages 1--19, Vancouver,
  Canada. Association for Computational Linguistics.

\bibitem[{Zhang et~al.(2020)Zhang, Zhao, Saleh, and Liu}]{zhang2020pegasus}
Jingqing Zhang, Yao Zhao, Mohammad Saleh, and Peter Liu. 2020.
\newblock Pegasus: Pre-training with extracted gap-sentences for abstractive
  summarization.
\newblock In \emph{International Conference on Machine Learning}, pages
  11328--11339. PMLR.

\bibitem[{Zhang et~al.(2019)Zhang, Wei, and Zhou}]{zhang-etal-2019-hibert}
Xingxing Zhang, Furu Wei, and Ming Zhou. 2019.
\newblock \href {https://doi.org/10.18653/v1/P19-1499} {{HIBERT}: Document
  level pre-training of hierarchical bidirectional transformers for document
  summarization}.
\newblock In \emph{Proceedings of the 57th Annual Meeting of the Association
  for Computational Linguistics}, pages 5059--5069, Florence, Italy.
  Association for Computational Linguistics.

\end{thebibliography}
